\documentclass[
	a4paper, 
	10pt, 
	unnumberedsections, 
	twoside, 
]{LTJournalArticle}
\usepackage{amsmath,amsfonts}
\usepackage[linesnumbered,ruled,vlined]{algorithm2e}
\usepackage{algorithmic}
\usepackage{array}
\usepackage{subfigure}
\usepackage{textcomp}
\usepackage{stfloats}
\usepackage{url}
\usepackage{verbatim}
\usepackage{graphicx}
\usepackage{hyperref}
\usepackage{multirow}
\usepackage{makecell}

\runninghead{LOSSGAT} 

\title{LOSS-GAT: Label Propagation and One-Class Semi-Supervised Graph Attention Network for Fake News Detection} 

\author{%
	Batool Lakzaei\textsuperscript{1}\thanks{Email: \href{mailto:b\_lakzaei@aut.ac.ir}{b\_lakzaei@aut.ac.ir}},
	Mostafa Haghir Chehreghani\textsuperscript{1}\thanks{Corresponding author. Email: \href{mailto:mostafa.chehreghani@aut.ac.ir}{mostafa.chehreghani@aut.ac.ir}} and 
	Alireza Bagheri\textsuperscript{1}\thanks{Email: \href{mailto:ar\_bagheri@aut.ac.ir}{ar\_bagheri@aut.ac.ir}}
}

\date{\footnotesize\textsuperscript{\textbf{1}}Department of Computer Engineering,
	Amirkabir University
	of Technology (Tehran Polytechnic)\\
	Tehran, Iran}

\renewcommand{\maketitlehookd}{%
	\begin{abstract}
		\noindent 
		In the era of widespread social networks, the rapid dissemination of fake news has emerged as a significant threat, inflicting detrimental consequences across various dimensions of people's lives.
		This apprehension has intensified in recent years, elevating the imperative for automated systems to detect and combat the proliferation of disinformation.
		Machine learning and deep learning approaches have been extensively employed for identifying fake news.
		However, a significant challenge in identifying fake news is the limited availability of
		labeled news datasets.
		Therefore, the One-Class Learning (OCL) approach, utilizing only a small set of labeled data from the interest class, can be a suitable approach to address this challenge.
		On the other hand, representing data as a graph enables access to diverse content and structural information, and label propagation methods on graphs can be effective in predicting node labels.
		In this paper, we adopt a graph-based model for data representation and introduce a semi-supervised and one-class approach for fake news detection, called LOSS-GAT.
		Initially, we employ a two-step label propagation algorithm, utilizing Graph Neural Networks (GNNs) as an initial classifier to categorize news into two groups: interest (fake) and non-interest (real). 
		Subsequently, we
		enhance the graph structure using structural augmentation techniques.
		Ultimately, we predict the final labels for all unlabeled data using a GNN that induces randomness within the local neighborhood of nodes through the aggregation function.
		We evaluate our proposed method on five common datasets and compare the results 
		against a set of baseline models, including both OCL and binary labeled models.
		The results demonstrate that LOSS-GAT achieves a notable improvement,
		surpassing 10\%, with the advantage of utilizing only a limited set of labeled fake news. 
		Noteworthy, LOSS-GAT even outperforms binary labeled models.
	\end{abstract}
}

\begin{document}

\maketitle 

\section{Introduction}
\label{sec:intro}

Today, the Internet and social networks are widely regarded as ideal tools for communication and information dissemination.
So, social networks have become the primary medium for interactions in today's societies \cite{paraschiv2022unified}.
The ease of widely disseminating information to millions of people within a short time has transformed social networks into a highly suitable platform for rapid information and news dissemination, particularly during critical situations such as natural disasters \cite{shu2020combating}.
Indeed, major social networks such as Twitter\footnote{\url{https://twitter.com/}} and Facebook\footnote{\url{https://www.facebook.com/}} have emerged as the primary sources of information on a global scale \cite{paraschiv2022unified}.

However, the absence of control and verification mechanisms for the accuracy of published information has led to social networks being deemed conducive to the rapid and extensive spread of disinformation \cite{zubiaga2018detection}.
Disinformation refers to false information deliberately published with the intent of deceiving and misleading the audience \cite{shu2020combating}.
Disinformation takes on various forms, such as fake news, rumors, hoax and clickbait.
Considering that the dissemination of disinformation can have severe negative effects on various aspects of human life, such as social \cite{takayasu2015rumor, gupta2013faking}, political \cite{allcott2017social}, economic \cite{carvalho2011persistent}, verifying the accuracy of information published on social networks is considered a critical issue in all societies.

In recent years, machine learning and deep learning algorithms have gained extensive utilization in detecting disinformation.
Most of the methods available in the research literature use a supervised classification approach to detect and classify disinformation.
However, the lack of access to a substantial and adequate set of labeled data is one of the fundamental challenges in the supervised approach. 
Because labeling extensive sets of information incurs significant human expenses.
Furthermore, the disinformation detection problem (e.g., fake news detection)
is intrinsically imbalanced, indicating that true information (e.g., real news) constitutes a larger proportion within the real world \cite{golo2023one}.

Generally, to overcome these challenges, various solutions have been proposed.
One-Class Learning (OCL), considered as a form of semi-supervised learning, is among the proposed solutions for this challenge.
One-Class Learning is a learning algorithm that exclusively employs labeled examples of the interest class (positive class) as its input \cite{bellinger2017sampling}.
Its primary goal is to construct a proficient model that effectively classifies new examples into interest (positive) or non-interest (negative) classes, accomplished through the strategic utilization of interest class examples.
Through instructing the classification model exclusively with examples from the interest class, OCL diminishes the gap in unbalanced training sets and eliminates the need to label examples of non-interest classes \cite{alam2020one}.
This approach is commonly employed in various domains, including fraud detection, anomaly detection and threat detection.
Given that the disinformation detection problem can be regarded as a form of anomaly detection, the One-Class Learning approach can be highly effective in addressing this issue.

One of the influential factors in the performance of learning algorithms is the data representation model.
Using an inappropriate model can disrupt the classification algorithm, especially when dealing with unlabeled data \cite{van2020survey}.
On the other hand, while OCL approach diminishes the need for labeled examples of non-interest classes, the learning process becomes more demanding owing to the absence of non-interest class examples for model training \cite{golo2023one}.
Hence, disinformation detection through OCL requires more robust  representation models to effectively separate false and true information in the hidden space \cite{golo2023one}.
Graph-based modeling is a powerful and efficient approach that facilitates the effective analysis of complex data structures in non-Euclidean space.
Indeed, graph-based modeling allows extracting class patterns that are challenging to discern using vector space models \cite{breve2011particle}.
Research results demonstrate that graph-based modeling can be highly advantageous for semi-supervised learning \cite{rossi2016optimization}.
Because graphs, with their ability to depict structural and contextual dependencies between nodes where each node represents an object, prove to be valuable in scenarios where the number of labels is limited \cite{shi2020skeleton}.
Typically, the process of label learning for unlabeled nodes in a graph is accomplished through a label propagation algorithm, wherein labeled nodes propagate their labels to  unlabeled nodes \cite{li2019lightweight}.
On the other hand, Graph Neural Networks (GNNs) stand as the most powerful tools for processing graph data.
In Graph Neural Networks, unlike traditional neural networks that rely on matrix representation, graph processing involves a series of simple and efficient computations, such as aggregating the local neighborhood of each node \cite{haghir2022half}.
The literature demonstrates that GNNs exhibit high efficiency in
semi-supervised learning \cite{kipf2016semi}.

In this paper, we propose the LOSS-GAT (Label Propagation and One-Class Semi-Supervised Learning with Graph Attention Network) algorithm, a novel One-Class Learning method for detecting fake news, utilizing GNNs along with a two-step label propagation approach.
We initiate the process by representing the data as a graph, comprising a limited number of labeled data (positive class = fake news) and a large number of unlabeled data (including positive class = fake news, and negative class = real news).
The methodology for constructing this graph is explained in detail in \autoref{sec:graph_construction}.
Then, we employ a two-step label propagation method to infer pseudo-labels for a subset of unlabeled data.
In the first step, we utilize a label propagation algorithm based on the Katz index, as described in \cite{de2022network}.
In the second step, we employ the pseudo-labels obtained from the first step to train a Graph Attention Network (GAT) model.
This model predicts the labels of unlabeled nodes and classifies them as either fake or real.
Then, a portion of these predicted labeled data is randomly selected as new pseudo-labels.
Finally, these pseudo-labels are utilized by the classification module.
In the classification module, we initiate the process by performing structural augmentation on the graph using the Adamic-Adar score \cite{adamic2003friends}.
After that, a GAT model is utilized to learn all nodes' embedding vectors, and the final labels are determined using a fully connected perceptron.
In this GAT network, we induce randomness to reconstruct neighborhoods and enhance the aggregation process.
The key novelties of our proposed method are as follows:
\begin{itemize}
	\item Our method accomplishes semi-supervised binary classification of fake news using only a limited amount of labeled data from the fake class.
	
	\item Our
	proposed two-step label propagation method is the first of its kind to utilize an initial classifier for label propagation.
	
	\item For the first time, we introduce an OCL approach utilizing GNNs for detecting fake news.
	
	\item Our method is the first approach that employs a structural augmentation to enhance the graph structure in the context of detecting fake news.
	
	\item 
	Our method is the first fake news detection algorithm that
	employs a probabilistic random approach grounded in node similarity degrees to reconstruct the neighborhood during the aggregation process.
\end{itemize}

The rest of this article is organized as follows:
In \autoref{sec:related_work}, we present related work.
In \autoref{sec:problem_formulation}, we formally describe the studied problem.
In \autoref{sec:proposed_model}, we introduce our proposed approach.
In \autoref{sec:experiments}, we present our experimental evaluation and 
the obtained results.
Finally, the paper is concluded in \autoref{sec:conclusion}.

\section{Related Work}
\label{sec:related_work}
The pervasive challenge of fake news in the digital era has prompted substantial research into effective detection and mitigation strategies. Researchers have explored various approaches, including natural language processing, machine learning, network analysis, and social media mining. This section provides a summary of the literature, emphasizing semi-supervised methods, which hold promise for leveraging both labeled and unlabeled data in fake news detection.

Dong et al. \cite{dong2020two} introduced a two-path deep semi-supervised learning framework for fake news detection in social media, using shared Convolutional Neural Networks.
Their model collaboratively optimizes loss functions to enhance accuracy with sparse labeled data.
Konkobo et al. \cite{konkobo2020deep} proposed a semi-supervised model for early fake news detection, utilizing three neural networks and four key components, including opinion extraction, user credibility assessment, hierarchical clustering for user networks and a semi-supervised news classifier.
The DEFD-SSL method \cite{al2023robust} employs diverse deep learners, augmentations, and distribution-aware pseudo-labeling with a hybrid loss function for balanced accuracy in imbalanced datasets
and benefits from ensemble learning and distribution alignment.
SLD-CNN \cite{mansouri2020semi} integrates CNNs with linear discriminant analysis, iteratively predicting labels for unlabeled data with an impact factor for effective error control.

Meel et al. \cite{meel2021temporal} proposed a semi-supervised CNN framework for fake news classification, leveraging linguistic and stylometric information with self-ensembling and Temporal Ensembling to enhance the accuracy.
Cross-SEAN \cite{paka2021cross} is a neural attention model that combines cross-stitch units with semi-supervised learning, integrating unlabeled data, external knowledge, tweet and user features, and adversarial training for both supervised and unsupervised loss.
Li et al. \cite{li2022novel} presented a self-learning semi-supervised deep learning network for detecting social media fake news. Featuring a confidence layer, it autonomously boosts accuracy by incorporating accurate results, proving advantageous for incomplete or limited datasets.
Yang \cite{yang2021covid} introduced an approach for COVID-19-related fake news on social media, using a text graph constructed from tweet similarity. The method employs a graph neural network   to classify fake news, reducing the need for extensive labeled data.

Hu et al. \cite{hu2019multi} introduced M-GCN, a framework for detecting fake news in short articles. It employs graph embedding, multi-depth GCN blocks, and an attention mechanism to combine multi-scale features, treating each news article as a node and leveraging speaker profiles for relationship networks.
Ren et al. \cite{ren2020adversarial} presented AA-HGNN, a model using a hierarchical attention mechanism in a heterogeneous information network with active learning for improved performance in data-scarce scenarios. It uses a two-level attention mechanism in a classifier and a selector, working together adversarially to enhance label quality and candidate selection.
Benamira et al. \cite{benamira2019semi} embedded articles, constructed an article similarity graph using a k-nearest-neighbors approach and employed Graph Convolutional Networks  and Graph Attention Networks for classification on the similarity graph.
Faustini and Covoes \cite{faustini2019fake} employed an OCL approach, training their model exclusively on fake news data. They introduced DCDistanceOCC, a novel algorithm highlighting OCL's effectiveness in discerning fake news across various text types.
G{\^o}lo et al. \cite{golo2023one} introduced a multimodal OCL approach
for fake news detection, utilizing Multimodal Variational Autoencoders (MVAEs) that learn from language, linguistic, and topic features.

In \cite{guacho2018semi}, tensor decomposition creates embeddings for news articles, forming a graph where known labels are propagated, transforming fake news detection into a semi-supervised problem. 
Xie et al. \cite{xie2022label} introduced LNMT, a Label Noise-Resistant Mean Teaching approach for weakly supervised fake news detection. It uses unlabeled news data and user feedback comments to generate improved weak labels, employing a two-stage training process with a baseline model to reduce noise impact.
De Souza et al. \cite{de2022network} introduced a fake news detection approach using Positive and Unlabeled Learning by Label Propagation (PU-LP), a transductive, semi-supervised algorithm adapted for text classification. They proposed network representations incorporating news objects and key n-grams to enhance the model, aiming to minimize labeling efforts, improve text classification accuracy, and leverage network structures for effective information extraction from unlabeled documents in fake news detection.

Our approach builds upon the foundation established in \cite{de2022network} by de Souza et al. We integrate three influential techniques: Graph Neural Networks, One-Class Learning, and Label Propagation. This synergy harnesses their strengths to effectively tackle the intricate challenges associated with the spread of fake news in the digital era.

\section{Problem Formulation}
\label{sec:problem_formulation}
Let 
$D = \{d_1, d_2, . . . , d_N\}$
be the news set with $N$ news items, where each news $d_i$ contains text content.
We define fake news detection as a one-class binary classification task utilizing a semi-supervised approach.
The class labels are represented by $y\in\{0, 1\}$, where $y = 1$ indicates fake news, and $y = 0$ indicates real news.
Following the semi-supervised and OCL approach, merely a small subset of news, denoted as $D_L \subset D,$ bears the label of fake news $(y=1)$.
The greater portion of the news dataset remains unlabeled $(D_U = D - D_L)$.
Our goal is to learn the model $F$ for predicting labels for all unlabeled data $D_U$, utilizing examples from the interest class (fake news):
\begin{equation}
	\label{eq:F}
	\hat{y} = F(d_i)
\end{equation}
where
$\hat{y} \in \{0,1\}$ is the predicted label, and 
$d_i \in D_U$ is an unlabeled example.
We summarize frequently used notations in \autoref{tab:notations}.
\begin{table*}[!t]
	\centering
	\caption{Notations and their descriptions.}
	\label{tab:notations}
	\begin{tabular}{l l}
		\hline
		\textbf{Notation} & \textbf{Definition} \\ \hline
		
		$D = \{d_1, d_2, ..., d_N\}$		& set of news items \\ \hline
		
		$N$		& number of news items in $D$ \\ \hline		
		
		$D_L \subset D$		& set of labeled news items \\ \hline
		
		$D_U \subset D$		& set of unlabeled news items \\ \hline
		
		$y \in \{0,1\}$		& class label (1 = fake news, and 0 = real news) \\ \hline
		
		$\hat{y} \in \{0,1\}$		& predicted class label (1 = fake news, and 0 = real news) \\ \hline
		
		$X \in R^{N \times l_t}$		& textual feature vectors \\ \hline
		
		$l_t$		& length of textual feature vectors \\ \hline
		
		$G_{knn} = (V_{knn} , E_{knn})$		& similarity k-nearest neighbors graph \\ \hline
		
		$A_{knn}$		& adjacency matrix of $G_{knn}$ \\ \hline 
		
		$G_{sa}= (V_{sa} , E_{sa})$		&structural augmented graph \\ \hline
		
		$W_{katz}$		& similarity matrix calculated by Katz index \\ \hline
		
		$IP_{katz} \subset D_U$		& set of inferred fake news by label propagation (step 1) module \\ \hline
		
		$IN_{katz} \subset D_U$		& set of inferred real news by label propagation (step 1) module \\ \hline
		
		$IP \subset D_U$		& set of inferred fake news by label propagation (step 2) module \\ \hline
		
		$IN \subset D_U$		& set of inferred real news by label propagation (step 2) module \\ \hline
		
		$H^l \in R^{N \times d_l}$		 & all node embeddings in layer $l$ of GATv2 \\ \hline	
		
		$AA(v,u)$		& Adamic-Adar score for nodes $v$ and $u$ \\ \hline
		
		$T_{aa}$		& the threshold for adding new edges based on Adamic-Adar score \\ \hline
		
		$N_c(v)$		& content neighbors of node $v$ \\ \hline
		
		$N_s(v)$		& structural neighbors of node $v$ \\ \hline
		
		$N(v) = N_c(v) \cup N_s(v)$		& all neighbors of node $v$ \\ \hline
		
	\end{tabular}
\end{table*}

\section{Proposed Model}
\label{sec:proposed_model}

\begin{figure}[!t]
	\centering
	\includegraphics[scale=.5]{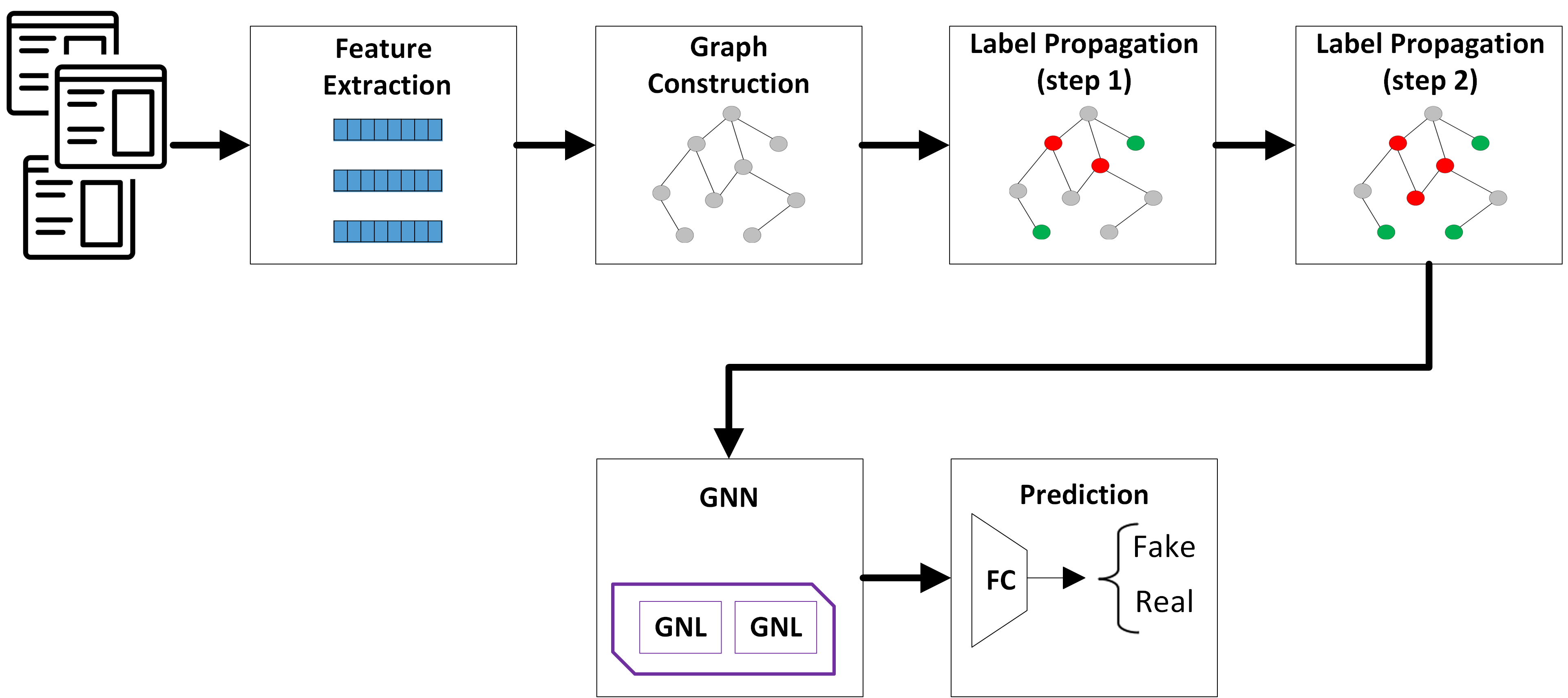}
	\caption{LOSS-GAT: our proposed approach for fake news detection}
	\label{fig:model}
\end{figure}

Our proposed model for fake news detection, illustrated in \autoref{fig:model},
consists of four main modules: feature extraction, graph construction, two-step label propagation, and classification.

Initially, the feature extraction module processes the textual content in news items using the Doc2Vec model \cite{le2014distributed} to learn their textual representation vectors.
Then, the graph construction module builds a similarity graph $G_{knn}$ among the news items.
In this graph, each news item is viewed as a node, and each node is connected to its $k$ nearest neighbors through undirected edges.
After that, the two-step label propagation module performs label propagation on $G_{knn}$ in two distinct steps.
In the initial step, label propagation is carried out utilizing the Katz index \cite{katz1953new}, leading to the inference of pseudo-labels for unlabeled nodes.
In the second step, the pseudo-labels inferred from the initial step are utilized, and a GNN is employed as the initial classifier to infer the final pseudo-labels for unlabeled nodes.
The classification module begins by applying structural augmentation to the $G_{knn}$ graph, constructing an augmented graph denoted as  $G_{sa}$.
Subsequently, a GNN is trained on the pseudo-labels inferred in the previous module to classify all unlabeled nodes within the $G_{sa}$ as either fake or real.
In this module, to enhance the GNN's performance, we propose inducing randomness.
Specifically, we propose selecting a randomized subset from the local neighborhood of each node during the aggregation process in GNN.
In the rest of this section, we describe each module in details.
\subsection{Feature extraction}
\label{sec:feature_extraction}

Findings in the literature demonstrate that fake news and real news exhibit distinct content-based features, including writing styles, lexical features, syntactic features, visual contents, and more \cite{shu2017fake, bondielli2019survey, meel2020fake}.
It can be asserted that the textual content stands as the most crucial aspect of any news item.
Hence, textual features rank among the most prevalent features employed for identifying fake news \cite{yu2021review}.
So, we generate initial feature vectors for news items through the analysis of their textual content.

Handling discrete words within textual content directly can pose a challenging task.
Therefore, textual content information is commonly embedded into a low-dimensional space \cite{wu2020rumor}.
The goal is to learn a mapping function 
$T: D \to X$,
which generates a text embedding vector for each news item according to its corresponding textual content.
A prevalent approach involves the utilization of the Word2Vec algorithm \cite{mikolov2013distributed}.
In this approach, individual words are represented as vectors and subsequently, the mean vector of all these words is adopted as the representation of the whole news article \cite{mikolov2013efficient}.
Our emphasis lies in detecting fake news disseminated through social networks, where there often exists a constraint on the word count within a single post.
As a result, the majority of news texts found on social networks are short texts.
The Word2Vec algorithm typically exhibits suboptimal performance within such scenarios \cite{wu2020rumor}.
Hence, we opt to utilize the Doc2Vec \cite{le2014distributed} approach,
an improved version of Word2Vec, to generate initial news embedding vectors:
\begin{equation}
	\label{eq:doc2vec}
	Doc2Vec(d_i) = x_i,
\end{equation}
where $d_i \in D$ denotes textual content for news item $i$,  
$x_i \in R^{l_t}$ denotes the text embedding vector for $d_i$, and 
$l_t$ denotes the length of text's embedding vectors.
\subsection{Graph construction}
\label{sec:graph_construction}

In the second module, we employ a graph-based modeling approach to structure news items.
Each news item is perceived as an individual node.
Subsequently, employing a distance metric such as the Euclidean distance, the similarity between text embeddings is computed.
Every node is linked to its $k$ nearest neighbors, building an undirected similarity graph $G_{knn} = (V_{knn}, E_{knn})$.
\subsection{Two-step label propagation}
\label{sec:two_step_lp}
The aim of the two-step label propagation module is to infer two sets: 
$IP \subset D_U$, which signifies interest news (fake news), and 
$IN \subset D_U$, representing non-interest news (real news).
Hence, $IP$ encompasses a subset of unlabeled news items that have been assigned a positive class (fake) label through the label propagation algorithm.
Meanwhile, $IN$ comprises a subset of unlabeled news items tagged with a negative class (real) label, inferred by the label propagation algorithm. 
To enhance the quality of the inferred pseudo-labels, thereby improving classification performance, we execute the label propagation process in two sequential steps.
During the initial step, we utilize the Katz index, while in the subsequent step, we employ an initial classifier.
\subsubsection{Step 1: label propagation using Katz index}
\label{sec:lp_1}
Within the $G_{knn}$ graph, edges are induced independently
and they all have the same weight $1$.
When two news items (nodes) are connected within this graph,
it depicts a high degree of similarity of their contents,
leading to a high likelihood that the nodes belong to the same class.
In the proposed $G_{knn}$ graph, only a small fraction of nodes bear a positive class label (fake news), while the majority of nodes remain unlabeled.
Labels can propagate along the pathways within this graph, implying that if two nodes share numerous common neighbors, there is a higher likelihood that they belong to the same class \cite{de2022network}.
There exist several methods for propagating labels in a graph.
In step 1, we utilize the Katz index, similar to the approach described by de Souza et al. \cite{de2022network}, for label propagation.
Katz
computes the similarity between node pairs based on all conceivable paths connecting them within the graph \cite{de2022network}. So: 
\begin{equation}	
	\label{eq:sim}
	\begin{split}
		sim(v_i , v_j) = \sum_{h=1}^{\infty} \alpha^h . |path^{<h>}_{v_i,v_j}| = 
		\\
		\alpha A_{knn}[i,j] + \alpha^2 A_{knn}^2[i,j] + \alpha^3  A_{knn}^3[i,j] + ... ,
	\end{split}	
\end{equation}
where 
$|path^{<h>}_{v_i,v_j}|$ is the number of paths of length $h$ from $v_i$ to $v_j$ and
$\alpha$ serves as a free parameter governing the impact of paths within the graph.
So, longer paths exert less influence compared to shorter ones \cite{de2022network}.
When $\alpha < 1/\epsilon$, where $\epsilon$ represents the largest eigenvalue of the matrix $A_{knn}$, \autoref{eq:sim} converges and can be computed using \autoref{eq:w}:
\begin{equation}
	\label{eq:w}
	W_{Katz} = (I - \alpha A_{knn})^{-1} - I.
\end{equation}
$W_{Katz} \in R^{N \times N}$,
and $I$ denotes the identity matrix.
We employ 
the Katz-based label propagation method of \cite{de2022network},
to infer two sets:
$IP_{Katz}$ (interest news = fake news) and $IN_{Katz}$ (non-interest news = real news), 
We utilize the similarity matrix $W_{Katz}$ and labeled news $D_L$ to compute these sets.
$IP_{Katz}$ encompasses news from $D_U$ that exhibit the highest similarity to examples in $D_L$. $IN_{Katz}$ comprises news from $D_U - IP_{Katz}$, signifying those news items that demonstrate the highest dissimilarity to the set $D_L \cup IP_{Katz}$.
\subsubsection{Step 2: Label propagation using an initial classifier}
\label{sec:lp_2}
In the second step, we apply an initial classifier for node-level classification, and predict pseudo-labels for all unlabeled nodes within the $G_{knn}$ graph.
Our proposed initial classifier relies on Graph Neural Networks (GNNs).
GNNs can learn node embeddings by simultaneously encoding both local graph structure and node features \cite{kipf2016semi}.
Within GNNs, the objective is to compute a function that maps each graph node to a vector in a low-dimensional vector space.
This mapping should be similarity preserving.
It means when two nodes share similar features and play similar roles in the graph's structure, they should be positioned closely to each other in the vector space.
The resulting vectors are referred to as node embedding vectors \cite{haghir2022half}.
In the course of an iterative process, these networks continuously update node embedding vectors as they traverse various layers.
Within each layer, an aggregation function is executed on the embeddings of each node and its neighboring nodes, subsequently updating the node embedding vector.
Various versions of Graph Neural Networks are available, and we propose utilizing the GATv2 network \cite{brody2021attentive}, which is an enhanced version of GAT \cite{velivckovic2017graph}.

The fundamental concept behind GAT (and GATv2) is that different neighbors of a node play distinct roles and hold varying levels of significance. Therefore, they contribute to the generation of the node's representation vector with distinct weights.
This network employs the multi-head attention mechanism to aggregate embedding vectors from the neighbors of each node \cite{velivckovic2017graph}.
The following equations delineate the computation of embedding vectors for each node across various layers in the GATv2 network \cite{brody2021attentive}:
\begin{equation}
	\label{eq:gat_1}
	h_v^l = \parallel_{k=1}^{K_{att}} \sigma (\sum_{u \in N(v)} \alpha_{vu}^k W^{l-1} h_u^{l-1})
\end{equation}
and
\begin{equation}
	\label{eq:gat_2}
	\alpha_{vu} = a \, \sigma (W^{l-1} h_v^{l-1} , W^{l-1} h_u^{l-1}),
\end{equation}
where $h_v^l$ represents the embedding of node $v$ in layer $l$,
$\parallel$ is the concatenation operator,
$K_{att}$ is the number of attention heads,
$\sigma$ is a non-linear function,
$N(v)$ consists of all neighbors of node $v$,
and $\alpha_{vu}$ is the attention coefficient between two nodes $v$ and $u$.
The weight matrix $W$ and the parameter $a$ are learned during the training process.
In summary, we utilize the following equation to represent $H^l \in R^{N \times d_l}$, which encompasses all node embeddings in layer $l$ ($d_l$ denotes the length of node embeddings):
\begin{equation}
	\label{eq:H}
	H^l = GATv2(G),
\end{equation}
where $G$ represents the input graph (e.g., $G_{knn}$).
We employ an L-Layer GATv2 network.
The embeddings produced in the final layer (layer L) are regarded as the ultimate embeddings. 
As previously mentioned, these embeddings can be employed as nodes feature vectors by various machine learning algorithms.
Hence, we feed these embedding vectors into a fully connected layer, ultimately predicting the final pseudo-labels for unlabeled nodes.
To train this network, we utilize the pseudo-labels generated in step 1 as the training set.
This module infers pseudo-labels for all unlabeled nodes in $D_U$, and classifies them as either fake news or real news.
Ultimately, we select a random portion of this classified data from each class to serve as the final pseudo-labels.
Algorithm 1 presents details of the process.
The final node embeddings $H^L$ are fed into a fully connected layer $FC$
to predict labels $\hat{Y}$.
After training, the unlabeled news set $D_U$ is classified into fake set $D_{fake}$
and real set $D_{real}$, based on these predicted labels.
The algorithm further employs the function $randomSample(D, p)$ to randomly select a specified percentage $p$ of news articles from each category,
forming the sets of inferred interest $IP$ and inferred non-interest $IN$. 
We use the following loss function to train the graph neural network of this module:
\begin{equation}
	\label{eq:cost_func_1}
	L(\Theta) = -\frac{1}{|D_{L'}|} \sum_{i=1}^{|D_{L'}|} y_{i} \, log(\hat{y}_{i}) \, 
	+ \,
	(1-y_{i}) \,
	log(1- \hat{y}_{i}),
\end{equation}
where $D_{L'} = D_L \cup IP_{Katz} \cup IN_{Katz}$,
$y_i$ is the ground truth label for $d_i$,
$\hat{y}_{i}$ is the predicted label for $d_i$, and $\Theta$ represents
the set of learnable parameters.

We assume that 
GNNs can approximate pseudo-labels with a high degree of accuracy due to the following reasons:
\begin{itemize}
	\item 
	The primary objective of label propagation is to identify the optimal method for propagating each node’s label to its neighboring nodes.
	One of the challenges in label propagation algorithms on graphs is the limited number of labeled nodes.
	Consequently, both node features and the local neighborhood play pivotal roles in the label propagation process.	
	\item
	GNNs are the most adept tools for graph processing, proficiently generating node embeddings through concurrent encoding of both the graph's local structure and the nodes' feature vectors.	
	\item
	The outcomes stemming from GNN application in various semi-supervised learning scenarios affirm that these models consistently attain impressive accuracy and often outperform numerous existing approaches.	
	
\end{itemize}
\begin{algorithm}[bht!]
	\label{alg:lp_2}
	\caption{Label propagation using an initial classifier (step 2)}
	
	\SetKwInput{KwInput}{Input}                
	\SetKwInput{KwOutput}{Output}              
	\DontPrintSemicolon
	
	\KwInput{\\
		$D_L$ : a set of labeled interest news (fake news)\\
		$D_U$ : a set of unlabeled news\\
		$IP_{Katz}$ : a set of interest news inferred in step 1 \\
		$IN_{Katz}$ : a set of non-interest news inferred in step 1\\
		$G_{knn} = (V_{knn}, E_{knn})$ : similarity graph \\
		$X \in R^{N \times l_t}$ : text embedding vectors\\
		$max\_epoch$ : number of epochs\\		
		$p$ :  selection percentage for pseudo-labels
	}
	\BlankLine
	
	\KwOutput{\\
		$IP$ : a set of inferred interest news (fake news) \\
		$IN$ : a set of inferred non-interest news (real news) 
	}
	\BlankLine
	\BlankLine
	\BlankLine
	
	$D_{L'} = D_L \cup IP_{Katz} \cup IN_{Katz}$\\	
	\BlankLine
	
	\tcp{Train $GATv2$ on $G_{knn}$}
	\For{$epoch: 1..max\_epoch$}
	{
		$H^0 = X$\\
		\For{$layer: 1..L$}
		{
			$H^{layer} = GATv2(G_{knn})$
		}
		\BlankLine
		$\hat{Y} = FC(H^L)$\\
		$loss \leftarrow \autoref{eq:cost_func_1}$\\
		$loss.backward()$		
	}
	\BlankLine
	
	$D_{fake} = \{\}$\\
	$D_{real} = \{\}$\\
	\ForEach{$d_i \in D_U$}
	{
		\If{$\hat{y}_i = 1$}
		{
			$D_{fake} = D_{fake} \cup\{d_i\}$
		}
		\BlankLine
		\If{$\hat{y}_i = 0$}
		{
			$D_{real} = D_{real} \cup\{d_i\}$
		}
	}
	
	\BlankLine
	$IP = randomSample(D_{fake},p)$\\
	$IN = randomSsample(D_{real},p)$	
	
\end{algorithm}

\subsection{Classification}
\label{sec:classification}
We perform node-level classification utilizing GNNs to predict the final labels for all unlabeled nodes.
To improve the performance of GNNs, we propose two enhancements:
1) structural augmentation of $G_{knn}$, and 
2) inducing randomness in the local neighborhood of nodes.

\subsubsection{Structural augmentation}
\label{sec:SA}

\begin{figure}[bht!]
	\centering
	\includegraphics[scale=.3]{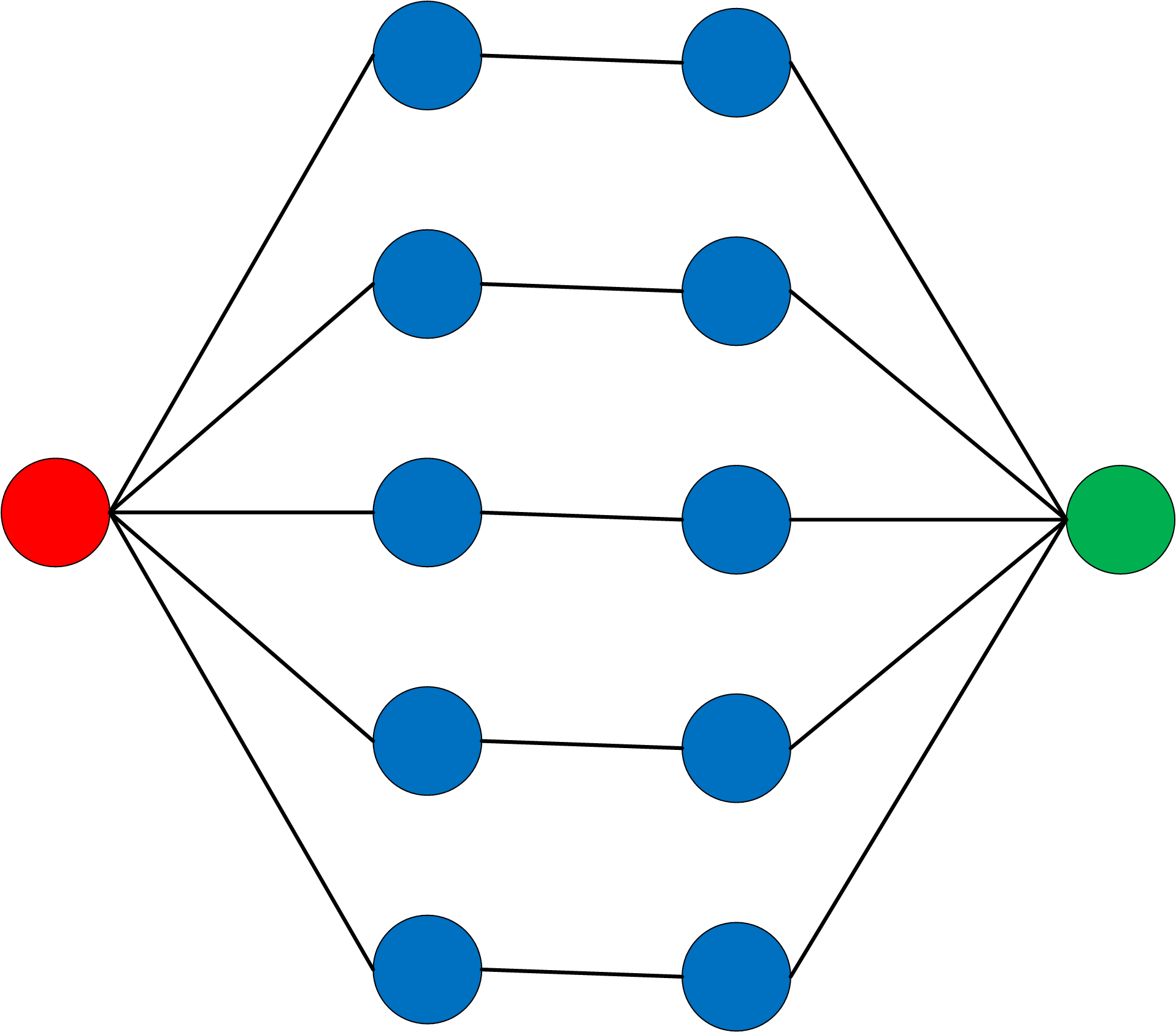}
	\caption{Structural similarity: the red and green nodes are not situated within the $2$-order neighborhood of each other.
		Therefore, if a two-layer GNN is employed, these nodes will not contribute to their embedding vectors.
		However, based on the graph structure, these nodes exhibit a high degree of structural similarity: in an isomorphism from the graph to itself (automorphism),
		these two nodes can be mapped to each other.}
	\label{fig:sa_sample}
\end{figure}

In an $L$-layer GNN, a computational graph with a depth of $L$ is generated for each node.
Consequently, neighbors whose maximum distance from the target node equals L exert an influence on the generation of the target node's embedding.
In the graph construction module, we modeled the dataset as a similarity graph $G_{knn}$, with edges determined by the similarity of news textual contents.
It is possible for two nodes within this graph to not be within each other's $L$-order neighborhood, yet they exhibit structural similarity.
An example is depicted in \autoref{fig:sa_sample}.
Hence, we propose the following form of structural augmentation:
\begin{itemize}
	\item 
	The structural similarity between nodes is computed using the Adamic-Adar (AA) \cite{adamic2003friends} criterion:
	\begin{equation}
		\label{eq:aa}
		AA(u,v) = \sum_{z \in N(u) \cap N(v)} \frac{1}{log |N(z)|},
	\end{equation}
	where $N(v)$ denotes the neighbors of node $v$.
	\item
	When the structural similarity between two nodes exceeds a threshold value, these nodes are deemed neighbors.
	Then, a weighted edge with $weight = AA(u,v)$
	is established between these two nodes.
	\begin{itemize}
		\item 
		If an edge already exists in the $G_{knn}$ between these two nodes, the AA index is added to the weight of this edge, i.e., $weight = 1+  AA(u,v)$.
	\end{itemize}
	\item
	The final local neighborhood of node $v$, denoted as $N(v)$, encompasses both the content neighborhood $N_c(v)$, that includes nodes within the proximity of $v$ in the similarity graph $G_{knn}$, and the structural neighborhood $N_s(v)$,
	that includes nodes which share structural similarity with $v$, i.e.,:
	\begin{equation}
		\label{eq:N}
		N(v) = N_c(v) \cup N_s(v).
	\end{equation}	
\end{itemize}
Algorithm 2 provides a detailed  description of
the structural augmentation process.
The algorithm initializes a structural augmented graph $G_{sa}$ with the same vertices as $G_{knn}$ (i.e., $V_{sa} = V_{knn}$) and an empty set of edges $E_{sa}$.
For each pair of vertices in $V_{sa}$, it computes the Adamic-Adar similarity using \autoref{eq:aa}.
If the computed similarity exceeds the given threshold $T_{aa}$, a weighted structural edge is added to $E_{sa}$.
\begin{algorithm}[bht!]
	\label{alg:sa}
	\caption{Structural augmentation.}
	
	\SetKwInput{KwInput}{Input}                
	\SetKwInput{KwOutput}{Output}              
	\DontPrintSemicolon
	
	\KwInput{\\		
		$G_{knn} = (V_{knn}, E_{knn})$: content similarity graph \\		
		$T_{aa}$:  Adamic-Adar threshold
	}
	\BlankLine
	
	\KwOutput{\\
		$G_{sa} = (V_{sa}, E_{sa})$: structural augmented graph		
	}
	\BlankLine
	\BlankLine
	\BlankLine
	
	$V_{sa} = V_{knn}$\\
	$E_{sa} = \{\}$\\
	\BlankLine
	
	\ForEach{$v \in V_{sa}$}
	{	\ForEach{$u \in V_{sa}$}
		{
			Compute $AA(v,u)$ using \autoref{eq:aa}\\
			\BlankLine
			
			\If{$AA(v,u) > T_{aa}$}
			{
				\If{$(v,u) \in E_{knn}$}
				{
					$E_{sa} = E_{sa} \cup \{(v,u, 1+ AA(v,u))\}$
				}
				\Else
				{
					$E_{sa} = E_{sa} \cup \{(v,u, AA(v,u))\}$	
				}
			}
			\ElseIf{$(v,u) \in E_{knn}$}
			{
				$E_{sa} = E_{sa} \cup \{(v,u, 1)\}$
			}		
		}
	}
	
	\BlankLine
	\BlankLine
	$G_{sa} = (V_{sa} , E_{sa})$

\end{algorithm}

In practical scenarios, the constructed graph may not precisely mirror our intended model due to various factors such as noise.
This challenge can introduce a disparity between the observed graph and the true graph, potentially impacting the outcomes of node classification.
In the optimal scenario, if we can transform the observed graph into the true graph by introducing hypothetical edges (absent in the observed graph) and eliminating irrelevant or unimportant edges (present in the observed graph), this can lead to a substantial enhancement in node classification.
Our proposed structural augmentation primarily aims to identify edges in the true graph that are absent in the observed graph.
The Adamic-Adar score represents a form of structural similarity between nodes and essentially calculates the likelihood of creating a link between nodes.
Hence, when the Adamic-Adar score suggests a high probability of creating a link between two nodes, it implies that there likely exists an edge between these nodes in the true graph, or an edge will likely emerge in the near future.
Consequently, these nodes will be positioned within each other's neighborhoods.
Hence, we can regard these two nodes as neighboring nodes in the observed graph, and the mutual influence they exert on each other during the neighborhood aggregation process can contribute to the generation of superior embeddings.
Therefore, this can lead to enhanced node classification.

\subsubsection{Inducing randomness in local neighborhood aggregation}
\label{sec:randomness}

Findings in the literature indicate that employing a random approach to construct neighborhoods within the graph can be highly effective in enhancing the performance of GNNs.
We suggest addressing challenges such as potential noises, the presence of irrelevant edges, and mitigating overfitting errors by crafting a node's neighborhood through a random and probabilistic approach that corresponds to the content and structural similarity of the nodes.
So, during the aggregation phase in each training iteration (epoch), rather than including the entire local neighborhood of each node, we opt to choose a subset of the node's neighbors. 
This subset is selected randomly and in accordance with the neighborhood's weight within the $G{sa}$ graph.

\subsubsection{Prediction}
\label{sec:pred}

Ultimately, we employ an $L$-layer GATv2 network to acquire the final nodes embeddings.
Next, these embeddings traverse through a fully connected layer, leading to the determination of the final labels for all unlabeled nodes.
We employ the following cost function to train this network:
\begin{equation}
	\label{eq:cost_func_2}
	L(\Theta) = -\frac{1}{|D_{L"}|} \sum_{i=1}^{|D_{L"}|} y_{i} \, log(\hat{y}_{i}) \, 
	+ \,
	(1-y_{i}) \,
	log(1- \hat{y}_{i}),
\end{equation}
where $D_{L"} = D_L \cup IP \cup IN$,
$y_i$ is the ground truth label for $d_i$,
$\hat{y}_{i}$ is the predicted label for $d_i$, and $\Theta$ represents the learnable parameters.
Further details are elucidated in Algorithm 3. 
The algorithm initiates by amalgamating labeled and inferred news in $D_{L"}$ to form the training set.
Then, it undergoes a training phase where a $GATv2$ network is trained on the structural augmented graph $G_{sa}$ for a specified number of epochs.
During each epoch, utilizing the function $getAllNeighbors(v, G_{sa})$ to obtain the neighbors of each node, the algorithm proceeds to randomly select $K$ neighbors for each node through the function $SelectNeighborsRandomly(neighbors, K)$,
where the selection process is guided by edge weights.
The selected neighbors are used to create a new graph $G'$, and the $GATv2$ model processes this graph to generate node embeddings.
The final layer node embeddings, i.e., matrix $H^L$, 
is then fed into a fully connected layer $FC$ to predict labels $\hat{Y}$.
The training process is steered by the loss function of \autoref{eq:cost_func_2}
and backpropagation is employed to optimize the model.
The resulted  model is proficient in predicting labels for all unlabeled news articles.

\begin{algorithm}
	\label{alg:classification}
	\caption{Classification}
	
	\SetKwInput{KwInput}{Input}                
	\SetKwInput{KwOutput}{Output}              
	\DontPrintSemicolon
	
	\KwInput{\\
		$D_L$ : a set of labeled interest news (fake news)\\
		$D_U$ : a set of unlabeled news\\
		$IP$ : a set of inferred interest news by Algorithm 1 \\ 
		$IN$ : a set of inferred non-interest news by Algorithm 1 \\ 
		$G_{sa} = (V_{sa}, E_{sa})$ : structural augmented graph \\
		$X \in R^{N \times l_t}$ : text embedding vectors\\
		$max\_epoch$ : number of epochs\\	
		$K$ : number of neighbors
	}
	\BlankLine
	
	\KwOutput{\\
		$\hat{Y}$ : predicted labels for all unlabeled news \\		
	}
	\BlankLine
	\BlankLine
	\BlankLine
	
	$D_{L"} = D_L \cup IP \cup IN$\\	
	\BlankLine
	\BlankLine
	
	\tcp{Train $GATv2$ on $G_{sa}$}
	\For{$epoch: 1..max\_epoch$}
	{
		\BlankLine
		
		$V' = V_{sa}$\\
		$E' = \{\}$\\
		\BlankLine
		
		\tcp{select K neighbors randomly}
		\ForEach{$v \in V_{sa}$}
		{
			$neighbors = getAllNeighbors(v, G_{sa})$\\
			\BlankLine
			
			\tcp{random selection based on weights}
			$selected\_neighbors = randomSelect(neighbors,K)$	
			\BlankLine
			
			\ForEach{$n_v \in selected\_neighbors$}
			{
				$E' = E' \cup \{(v,n_v)\}$
			}	
		}
		
		$G' = (V', E') $\\
		
		\BlankLine
		\BlankLine
		
		$H^0 = X$\\
		\For{$layer: 1..L$}
		{
			$H^{layer} = GATv2(G')$
		}
		\BlankLine
		$\hat{Y} = FC(H^L)$\\
		$loss \leftarrow \autoref{eq:cost_func_2}$\\
		$loss.backward()$		
	}
	\BlankLine
	\BlankLine
	\BlankLine
	$H^{L} = GATv2(G')$\\
	$\hat{Y} = FC(H^L)$\\

\end{algorithm}

\section{Experiments}
\label{sec:experiments}
In this section, first we provide an overview of the datasets, the baseline models and the evaluation metrics used in our empirical evaluations. 
Then, we report the results of our extensive experiments,
encompassing diverse settings and conditions to comprehensively assess our approach \footnote{The code of our method is publicly available at:
	\url{https://github.com/blakzaei/LOSS_GAT} .}.

\subsection{Datasets}
\label{sec:datasets}
We utilize five fully labeled datasets to validate our proposed approach—three in English and two in Portuguese: Fake News Net, Fake.Br, Fact-checked News, Fake News Detection, and Fake News Data.
The first dataset, FakeNewsNet (FNN), is sourced from the FakeNewsNet repository\footnote{\url{https://github.com/KaiDMML/FakeNewsNet}} \cite{shu2020fakenewsnet} and comprises English news articles featuring famous people,
fact-checked by the GossipCop\footnote{\url{https://www.gossipcop.com/}} website.
The dataset encompasses 21,032 news articles.
Nevertheless, an initial analysis reveals a notable imbalance in the distribution of tokens within these articles.
So, to ensure uniformity in dataset sizes and mitigate potential classification bias, we opt for news articles that fall within the range of $200$ to $600$ tokens.
This range is determined by considering the words present in the news after removing extraneous characters and stopwords.
Consequently, FNN comprises 1,705 fake news articles and 5,298 real news articles, making it the dataset with the most pronounced class imbalance.

The second dataset, Fake.Br\footnote{\url{ https://github.com/roneysco/Fake.BR-Corpus}}, is the first dataset in Portuguese created for fake news detection.
This dataset is meticulously collected and labeled by human annotators.
It encompasses a total of 7,200 news articles, evenly distributed with 3,600 being real news and 3,600 being fake news. 
These articles span across six distinct categories: politics (58\%), TV and celebrities (21.4\%), society and daily life (17.7\%), science and technology (1.5\%), economy (0.7\%), and religion (0.7\%) \cite{silva2020towards}.
All news items are accessible in two formats: complete and truncated.
Text truncation is employed to create a dataset with a consistent word count across all texts.
We opt for truncated news articles to prevent any bias from affecting the learning process.

The third dataset, Fact-checked News\footnote{\url{https://github.com/GoloMarcos/FKTC}},
is meticulously curated by aggregating fact-checking reports from reputable sources such as AosFatos\footnote{\url{https://aosfatos.org/noticias/}},
Agência Lupa\footnote{\url{https://piaui.folha.uol.com.br/lupa/}},
Fato ou Fake\footnote{\url{https://g1.globo.com/fato-ou-fake/}},
UOL Confere\footnote{\url{https://noticias.uol.com.br/confere/}},
and G1-Política\footnote{\url{https://g1.globo.com/politica/}}.
It comprises 2,168 Brazilian political news articles, with 1,124 being real and 1,044 being fake. These articles are gathered between August 2018 and May 2019.
Within this timeframe, Brazil conducted its presidential elections.
Leading up to the elections, a surge in disinformation circulated on social media platforms, paving the way for the development of this dataset.

The fourth dataset, Fake News Detection\footnote{\url{https://www.kaggle.com/jruvika/fake-news-detection}},
is uploaded to the Kaggle website by Jruvika.
It initially encompassed 4,009 news items. However, following an initial data cleaning phase, which involved the removal of entries with missing values, 3,988 news articles remain--omprising 2,121 Fake and 1,867 Real news items.

The fifth dataset, Fake News Data\footnote{\url{https://www.kaggle.com/c/fake-news/data}}, was originally hosted on Kaggle website in 2018 as part of a Kaggle competition. It has since been made openly accessible with annotations, serving as a valuable resource for analysis and learning purposes.
This dataset comprises a total of 20,800 samples. After eliminating entries with missing fields, there are 20,700 entries remaining, featuring a balanced distribution of 10,360 Fake news and 10,340 Real news entries.

\autoref{tab:ds} presents a summary of the characteristics of these datasets.
In the initial phase, we conduct preliminary preprocessing on the textual content of each of the mentioned datasets. 
This includes actions such as converting characters to lowercase, eliminating stopwords, removing links and numerical content, and reducing words to their stems using the PorterStemmer from the Natural Language Toolkit (NLTK).
Subsequently, we employ the Doc2Vec technique to transform each news article into a 500-dimensional feature vector.
\begin{table*}[!t]
	\centering
	\caption{Summary of the news datasets used in the experimental evaluation.}
	\label{tab:ds}
	\begin{tabular}{l l l l l}
		\hline
		\textbf{Dataset} & \textbf{Language} & \textbf{Total Entries} & \textbf{Fake News} & \textbf{Real News} \\ \hline
		FakeNewsNet (FNN)         & English    & 7003  & 1705  & 5298  \\ \hline
		Fake.Br             & Portuguese & 7200  & 3600  & 3600  \\ \hline
		Fact-checked News   & Portuguese & 2168  & 1044  & 1124  \\ \hline
		Fake News Detection & English    & 3988  & 2121  & 1867  \\ \hline
		Fake News Data      & English    & 20700 & 10360 & 10340 \\ \hline
	\end{tabular}
\end{table*}
\subsection{Baseline Models}
\label{sec:baselines}

We assess the performance of our proposed model by comparing it against seven semi-supervised fake news detection methods:
\begin{itemize}
	\item \textbf{PULP-LPHN \cite{de2022network}:} 
	An approach based on positive and unlabeled learning through label propagation in heterogeneous networks for fake news detection.
	Our proposed model extends upon this method.
	
	\item \textbf{OCSVM \cite{manevitz2001one}:}
	One-Class Support Vector Machine (OCSVM) is a machine learning algorithm employed for anomaly detection.
	It discerns rare data points from normal ones by identifying a hyperplane that segregates the bulk of the data from potential outliers.
	In our scenario, fake news constitutes the normal data points, while real news serves as the anomalies.
	
	\item \textbf{OCRF \cite{desir2012random}:}
	One-Class Random Forest (OCRF) is an extension of the traditional Random Forest algorithm which creates an ensemble of decision trees to identify outliers or rare instances (real news) in a dataset by learning the distribution of normal data (fake news).
	
	\item \textbf{KNND \cite{tan2019introduction}:}
	Anomaly Detection using K-Nearest Neighbors (KNND) is a technique that identifies real news in a dataset by measuring the similarity of a data point to its K nearest neighbors, where deviations from the majority of neighbors suggest anomalous behavior.
	
	\item \textbf{KMeans \cite{tan2019introduction}:}
	Anomaly Detection using KMeans is an approach that identifies real news in a dataset by clustering the data into K clusters and flagging data points that are significantly distant from their cluster centroids as potential real news.	
\end{itemize}
Furthermore, we conduct a performance comparison between our proposed method, which relies on a One-Class Learning approach utilizing solely the labeled data of the positive class (fake news) for binary classification, and semi-supervised methods that incorporate labels from both classes (positive and negative) in the training dataset: Binary Labeled Decision Tree (BLDT) and Binary Labeled Random Forest (BLRF) classifiers.
The training dataset for these two classifiers comprises labeled samples from both the fake and real classes.
To leverage the semi-supervised setting of the problem, we utilize the "sklearn.semi\_supervised.SelfTrainingClassifier"\footnote{\url{https://scikit-learn.org/stable/modules/semi\_supervised.html\#self-training}} from the scikit-learn library to train these models.

All of the aforementioned methods are implemented using the Python programming language.
The configuration and parameter settings for each of these methods are detailed in the \autoref{sec:setup}

\subsection{Evaluation Metrics}
\label{sec:metrics}
Numerous standard metrics are available to assess the performance of a binary classification model, with accuracy often regarded as one of the most appropriate criteria.
This metric is defined as follows:
\begin{equation}
	\label{eq:acc}
	Acc = \frac{(TP+TN)}{(TP+FP+FN+TN)}.
\end{equation}
Here, $TP$, $TN$, $FP$ and $FN$ indicate True Positive, True Negative, False Positive, and False Negative, respectively.

In scenarios where the data distribution across different classes is imbalanced, it becomes imprudent to assess methods solely based on accuracy.
This is due to the tendency for methods that classify new samples into the more abundant class to exhibit higher accuracy, thus potentially masking the true performance in distinguishing the minority class \cite{wu2020rumor}.
Given the nature of the fake news detection problem, which often involves imbalanced datasets, including the FNN dataset we utilize, we employ additional evaluation criteria such as macro-F1 and interest-F1 (F1 only for fake news as interest class).
The macro-F1 score for a dataset with k classes is computed through the following equations:
\begin{equation}
	P_i = \frac{TP_i}{TP_i + FP_i} \, \, \, R_i = \frac{TP_i}{TP_i + FN_i}
\end{equation}

\begin{equation}
	P_{Macro} = \frac{\sum_{i=1}^{k} P_i}{k} ,
	\, \, \,
	R_{Macro} = \frac{\sum_{i=1}^{k} R_i}{k}
\end{equation}
\begin{equation}
	Macro-F1 = \frac{2 P_{Macro} \times R_{Macro}}{P_{Macro} + R_{Macro}}
\end{equation}


\subsection{Experimental setup}
\label{sec:setup}
This section outlines the experimental setup for the baseline models and our proposed model.
Detailed parameters values for all models can be found in \autoref{tab:p_values}. 
\begin{table*}[!t]
	\centering
	\caption{Parameters values.}
	\label{tab:p_values}	
	\begin{tabular}{l l l}
		\hline
		\textbf{Model} &
		\textbf{Parameter} &
		\textbf{Value(s)} \\ \hline
		\multirow{4}{*}{\textbf{LOSS-GAT}} &
		\makecell[l]{k (number of nearest neighbors to build KNN graph} &
		$\{5, \, 6, \, 7\}$ \\ \cline{2-3} 
		&
		$\alpha$ (\autoref{eq:w}) &
		$\{0.005, \, 0.01, \, 0.02\}$ \\ \cline{2-3} 
		&
		\makecell[l]{$T_{aa}$ (threshold for  adding new \\edges based on Adamic-Adar score): Algorithm 2}& 
		$m \times 10^{-1} , m \in [1..6]$ \\ \cline{2-3} 
		&
		\makecell[l]{p (selection   percentage) for pseudo-labels: Algorithm 1}& 
		$m \times 10^{-1} , m \in [5..7]$ \\ \hline
		\multirow{4}{*}{\textbf{PULP-LPHN}} &
		convergence threshold &
		$0.00005$ \\ \cline{2-3} 
		&
		maximum number of iterations &
		$1000$ \\ \cline{2-3} 
		&
		\makecell[l]{k (number of nearest neighbors to build KNN graph)} &
		$\{5, \, 6, \, 7\}$ \\ \cline{2-3} 
		&
		$\alpha$ (\autoref{eq:w}) &
		$\{0.005, \, 0.01, \, 0.02\}$ \\ \hline
		\multirow{2}{*}{\textbf{OCSVM}} &
		$\gamma$ (kernel coefficient) &
		$1 \times 10^m , m \in [-3..1] $ \\ \cline{2-3} 
		&
		kernel &
		\{Linear, RBF\} \\ \hline
		\multirow{2}{*}{\textbf{OCRF}} &
		\makecell[l]{contamination (the proportion of outliers in the data set)} &
		$m \times 10^{-1} , m \in [1..5]$ \\ \cline{2-3} 
		&
		\makecell[l]{max\_features (the number of features to draw from X to \\ train each base estimator)} &
		$m \times 10^{-1} , m \in [1..9]$ \\ \hline
		\multirow{2}{*}{\textbf{KNND}} &
		k (the number of nearest neighbors) &
		$1 + 3 \times m , m \in [1..7]$ \\ \cline{2-3} 
		&
		anomaly threshold &
		$0.05 \times m , m \in [1..19]$ \\ \hline
		\multirow{2}{*}{\textbf{KMeans}} &
		k (the number of clusters) &
		$1 + 2 \times m , m \in [1..9]$ \\ \cline{2-3} 
		&
		anomaly threshold &
		$0.05 \times m , m \in [1..19]$ \\ \hline
		\multirow{3}{*}{\textbf{BLDT (self-train)}} &
		\makecell[l]{criterion (The selection criterion used to select which \\ labels to add to the training set)} &
		\makecell[l]{K\_best} \\ \cline{2-3} 
		&
		\makecell[l]{k\_best (The amount of samples to add in each iteration)} &
		$m \times 10, 10 \in [1..7]$ \\ \cline{2-3} 
		&
		\makecell[l]{max\_depth (the maximum depth of the tree)} &
		$\{4, \, 8, \, 10, \, 15, \, 20, \,50\}$ \\ \hline
		\multirow{4}{*}{\textbf{BLRF (self-train)}} &
		criterion &
		K\_best \\ \cline{2-3} 
		&
		k\_best &
		$m \times 10, 10 \in [1..7]$ \\ \cline{2-3} 
		&
		n\_estimators &
		$\{20, \,50, \, 70, \,100\}$ \\ \cline{2-3} 
		&
		max\_depts &
		$\{4,\, 8, \, 10, \, 15, \, 20, \,50\}$ \\ \hline
		
	\end{tabular}%
\end{table*}

To assess the model, we examine three distinct scenarios based on the quantity of labeled data, exclusively comprising fake labels: 10\%, 20\%, and 30\%.
For instance, within the 10\% scenario, merely 10\% of the dataset's fake news is designated as labeled data, while all other instances of fake news and every example of real news are categorized as unlabeled data. 
Each scenario is replicated 10 times across all algorithms.
During each replication, diverse data points are randomly selected to serve as labeled data.
Taking into account that for each algorithm, distinct values are assigned to its critical parameters (as indicated in the \autoref{tab:p_values}), the maximum value achieved (accuracy, macro-F1 and interest-F1) is regarded as the  outcome for the respective algorithm in each iteration.
Then, the average of the outcomes from 10 repetitions is deemed the final value for each algorithm.

\subsection{Results}
\label{sec:results}
In this section, we present the key findings of our experiments.
We demonstrate that OCL-based approaches can yield strong performance, surpassing even the binary labeled models in fake news detection.
Employing these models with only a small set of labeled fake news, while circumventing the need to label examples from real news, is viewed as a significant potential advantage that addresses the challenge of data labeling in this context.
Furthermore, we will demonstrate that our proposed model, incorporating GNNs and a two-step label propagation strategy, can outperform the other models under investigation, yielding superior results.
The findings suggest that introducing the second step in the label propagation process has a beneficial impact on enhancing the overall results.
The results from applying various methods to the five datasets, mentioned in \autoref{sec:datasets}, are displayed in \autoref{tab:results_fnn}, \autoref{tab:results_fakebr}, \autoref{tab:results_fact_checked}, \autoref{tab:results_fn_detection}, and \autoref{tab:results_fn_data} . 
Evaluation criteria include accuracy, interest-F1, and macro-F1 scores.
Our emphasis lies on the imbalanced dataset (FNN) concerning the macro-F1 score, while for the balanced dataset, we prioritize the accuracy score.
In each table, the percentages of fake news labeled for training the models are 10\%, 20\%, and 30\%. 
The best values in each table are highlighted in bold.
The results demonstrate that our proposed model can yield a substantial improvement (exceeding 10\%) in comparison with the other baseline models,
including those utilizing labeled data in both classes for training.

To enhance the depth  of our empirical analysis,
in the following we try to answer these experimental questions:

\begin{itemize}
	\item 
	Q1: To what extent does the two-step label propagation process contribute to enhancing the model's performance?
	\item 
	Q2: Does the quantity of labeled data significantly influence the final results?
	\item 
	Q3: Has the OCL approach demonstrated superior performance compared to binary labeled methods?
	\item
	Q4: Does LOSS-GAT outperform other baseline models for fake news detection?
\end{itemize}

\begin{table*}[bht!]
	\centering
	\tiny
	\caption{Performance comparison over FakeNewsNet (FNN)}
	\label{tab:results_fnn}
	
	\begin{tabular*}{\textwidth}{@{\extracolsep{\fill}}c ccc ccc ccc}
		\hline
		\multirow{2}{*}{\textbf{method}} &
		\multicolumn{3}{c}{\textbf{10\%}} &
		\multicolumn{3}{c}{\textbf{20\%}} &
		\multicolumn{3}{c}{\textbf{30\%}} \\ \cline{2-10} 
		&
		\multicolumn{1}{c}{\textbf{Acc}} &
		\multicolumn{1}{c}{\textbf{Intereset-F1}} &
		\textbf{Macro-F1} &
		\multicolumn{1}{c}{\textbf{Acc}} &
		\multicolumn{1}{c}{\textbf{Intereset-F1}} &
		\textbf{Macro-F1} &
		\multicolumn{1}{c}{\textbf{Acc}} &
		\multicolumn{1}{c}{\textbf{Intereset-F1}} &
		\textbf{Macro-F1} \\ \hline
		
		LOSS-GAT &
		\multicolumn{1}{c}{\textbf{0.8253$\pm$0.0019}} &			
		\multicolumn{1}{c}{\textbf{0.6116$\pm$0.0036}} &
		\textbf{0.7477$\pm$0.0023} &
		\multicolumn{1}{c}{\textbf{0.8365$\pm$0.0031}} &
		\multicolumn{1}{c}{\textbf{0.6034$\pm$0.0074}} &
		\textbf{0.7491$\pm$0.0043} &
		\multicolumn{1}{c}{\textbf{0.8447$\pm$0.0037}} &
		\multicolumn{1}{c}{\textbf{0.5868$\pm$0.0047}} &
		\textbf{0.7447$\pm$0.0015} \\ \hline
		
		PULP-LPHN &
		\multicolumn{1}{c}{0.7288$\pm$0.0145} &
		\multicolumn{1}{c}{0.4371$\pm$0.0169} &
		0.629$\pm$0.0069 &
		\multicolumn{1}{c}{0.7289$\pm$0.0111} &
		\multicolumn{1}{c}{0.4548$\pm$0.0097} &
		0.6371$\pm$0.008 &
		\multicolumn{1}{c}{0.7158$\pm$0.009} &
		\multicolumn{1}{c}{0.4441$\pm$0.01} &
		0.6266$\pm$0.008 \\ \hline
		
		OCSVM &
		\multicolumn{1}{c}{0.6336$\pm$0.0161} &
		\multicolumn{1}{c}{0.349$\pm$0.0092} &
		0.5468$\pm$0.0058 &
		\multicolumn{1}{c}{0.6239$\pm$0.0117} &
		\multicolumn{1}{c}{0.3345$\pm$0.0052} &
		0.5361$\pm$0.0039 &
		\multicolumn{1}{c}{0.6184$\pm$0.009} &
		\multicolumn{1}{c}{0.3155$\pm$0.0069} &
		0.5254$\pm$0.0039 \\ \hline
		
		OCRF &
		\multicolumn{1}{c}{0.6452$\pm$0.0163} &
		\multicolumn{1}{c}{0.3518$\pm$0.0086} &
		0.5536$\pm$0.0057 &
		\multicolumn{1}{c}{0.637$\pm$0.0094} &
		\multicolumn{1}{c}{0.3395$\pm$0.0062} &
		0.5446$\pm$0.0051 &
		\multicolumn{1}{c}{0.6339$\pm$0.0107} &
		\multicolumn{1}{c}{0.3185$\pm$0.0067} &
		0.5341$\pm$0.0051 \\ \hline
		
		KNND &
		\multicolumn{1}{c}{0.6028$\pm$0.0008} &
		\multicolumn{1}{c}{0.2376$\pm$0.0028} &
		0.4845$\pm$0.0016 &
		\multicolumn{1}{c}{0.6225$\pm$0.0229} &
		\multicolumn{1}{c}{0.2152$\pm$0.0186} &
		0.483$\pm$0.0023 &
		\multicolumn{1}{c}{0.6797$\pm$0.001} &
		\multicolumn{1}{c}{0.1623$\pm$0.0049} &
		0.4822$\pm$0.0027 \\ \hline
		
		KMeans &
		\multicolumn{1}{c}{0.6303$\pm$0.0127} &
		\multicolumn{1}{c}{0.2166$\pm$0.0106} &
		0.4872$\pm$0.0012 &
		\multicolumn{1}{c}{0.6645$\pm$0.0171} &
		\multicolumn{1}{c}{0.1879$\pm$0.0138} &
		0.4881$\pm$0.0017 &
		\multicolumn{1}{c}{0.6905$\pm$0.0104} &
		\multicolumn{1}{c}{0.1672$\pm$0.0108} &
		0.4885$\pm$0.0026 \\ \hline
		
		BLDT &
		\multicolumn{1}{c}{0.7286$\pm$0.0141} &
		\multicolumn{1}{c}{0.4423$\pm$0.0219} &
		0.6313$\pm$0.0098 &
		\multicolumn{1}{c}{0.7612$\pm$0.0135} &
		\multicolumn{1}{c}{0.4533$\pm$0.0129} &
		0.6502$\pm$0.0068 &
		\multicolumn{1}{c}{0.7674$\pm$0.0102} &
		\multicolumn{1}{c}{0.4603$\pm$0.0144} &
		0.656$\pm$0.0079 \\ \hline
		
		BLRF &
		\multicolumn{1}{c}{0.7638$\pm$0.0082} &
		\multicolumn{1}{c}{0.0588$\pm$0.055} &
		0.4619$\pm$0.0296 &
		\multicolumn{1}{c}{0.7582$\pm$0.0053} &
		\multicolumn{1}{c}{0.0228$\pm$0.0212} &
		0.4424$\pm$0.0121 &
		\multicolumn{1}{c}{0.7596$\pm$0.0036} &
		\multicolumn{1}{c}{0.0159$\pm$0.0066} &
		0.4395$\pm$0.0042 \\ \hline
	\end{tabular*}%
	
\end{table*}
\begin{table*}[bht!]
	\centering
	\tiny
	\caption{Performance comparison over Fake.Br}
	\label{tab:results_fakebr}
	\resizebox{\textwidth}{!}{%
		\begin{tabular*}{\textwidth}{@{\extracolsep{\fill}}c ccc ccc ccc}
			\hline
			\multirow{2}{*}{\textbf{method}} &
			\multicolumn{3}{c}{\textbf{10\%}} &
			\multicolumn{3}{c}{\textbf{20\%}} &
			\multicolumn{3}{c}{\textbf{30\%}} \\ \cline{2-10} 
			&
			\multicolumn{1}{c}{\textbf{Acc}} &
			\multicolumn{1}{c}{\textbf{Intereset-F1}} &
			\textbf{Macro-F1} &
			\multicolumn{1}{c}{\textbf{Acc}} &
			\multicolumn{1}{c}{\textbf{Intereset-F1}} &
			\textbf{Macro-F1} &
			\multicolumn{1}{c}{\textbf{Acc}} &
			\multicolumn{1}{c}{\textbf{Intereset-F1}} &
			\textbf{Macro-F1} \\ \hline
			LOSS-GAT &
			\multicolumn{1}{c}{\textbf{0.7412$\pm$0.0095}} &
			\multicolumn{1}{c}{\textbf{0.7211$\pm$0.0129}} &
			\textbf{0.7397$\pm$0.0097} &
			\multicolumn{1}{c}{\textbf{0.7641$\pm$0.0096}} &
			\multicolumn{1}{c}{\textbf{0.7339$\pm$0.008}} &
			\textbf{0.7609$\pm$0.0088} &
			\multicolumn{1}{c}{\textbf{0.7774$\pm$0.0037}} &
			\multicolumn{1}{c}{\textbf{0.7343$\pm$0.0029}} &
			\textbf{0.7708$\pm$0.0027} \\ \hline
			PULP-LPHN &
			\multicolumn{1}{c}{0.6335$\pm$0.0092} &
			\multicolumn{1}{c}{0.5938$\pm$0.0085} &
			0.63$\pm$0.0089 &
			\multicolumn{1}{c}{0.6823$\pm$0.0082} &
			\multicolumn{1}{c}{0.6332$\pm$0.0092} &
			0.6765$\pm$0.0082 &
			\multicolumn{1}{c}{0.7113$\pm$0.0054} &
			\multicolumn{1}{c}{0.6433$\pm$0.0077} &
			0.7004$\pm$0.0058 \\ \hline
			OCSVM &
			\multicolumn{1}{c}{0.5232$\pm$0.0029} &
			\multicolumn{1}{c}{0.4741$\pm$0.0145} &
			0.5188$\pm$0.0038 &
			\multicolumn{1}{c}{0.5254$\pm$0.0016} &
			\multicolumn{1}{c}{0.4758$\pm$0.0146} &
			0.5209$\pm$0.0023 &
			\multicolumn{1}{c}{0.5283$\pm$0.0025} &
			\multicolumn{1}{c}{0.4602$\pm$0.0056} &
			0.5207$\pm$0.0024 \\ \hline
			OCRF &
			\multicolumn{1}{c}{0.5267$\pm$0.0038} &
			\multicolumn{1}{c}{0.5069$\pm$0.0155} &
			0.5254$\pm$0.0033 &
			\multicolumn{1}{c}{0.5281$\pm$0.005} &
			\multicolumn{1}{c}{0.4885$\pm$0.0144} &
			0.5249$\pm$0.0043 &
			\multicolumn{1}{c}{0.5297$\pm$0.0042} &
			\multicolumn{1}{c}{0.4665$\pm$0.0066} &
			0.523$\pm$0.0042 \\ \hline
			KNND &
			\multicolumn{1}{c}{0.5068$\pm$0.0013} &
			\multicolumn{1}{c}{0.4933$\pm$0.002} &
			0.5065$\pm$0.0014 &
			\multicolumn{1}{c}{0.5074$\pm$0.0016} &
			\multicolumn{1}{c}{0.4782$\pm$0.0026} &
			0.5058$\pm$0.0017 &
			\multicolumn{1}{c}{0.5154$\pm$0.0108} &
			\multicolumn{1}{c}{0.44$\pm$0.0322} &
			0.5048$\pm$0.0017 \\ \hline
			KMeans &
			\multicolumn{1}{c}{0.489$\pm$0.0069} &
			\multicolumn{1}{c}{0.4217$\pm$0.0182} &
			0.4814$\pm$0.0033 &
			\multicolumn{1}{c}{0.4958$\pm$0.0116} &
			\multicolumn{1}{c}{0.4029$\pm$0.0244} &
			0.4822$\pm$0.0032 &
			\multicolumn{1}{c}{0.5125$\pm$0.0106} &
			\multicolumn{1}{c}{0.3656$\pm$0.0182} &
			0.4844$\pm$0.0028 \\ \hline
			BLDT &
			\multicolumn{1}{c}{0.679$\pm$0.0066} &
			\multicolumn{1}{c}{0.6774$\pm$0.0124} &
			0.6788$\pm$0.0065 &
			\multicolumn{1}{c}{0.7072$\pm$0.0082} &
			\multicolumn{1}{c}{0.6999$\pm$0.0092} &
			0.7067$\pm$0.0081 &
			\multicolumn{1}{c}{0.7083$\pm$0.0058} &
			\multicolumn{1}{c}{0.6969$\pm$0.0104} &
			0.7077$\pm$0.006 \\ \hline
			BLRF &
			\multicolumn{1}{c}{0.7301$\pm$0.0035} &
			\multicolumn{1}{c}{0.7258$\pm$0.0075} &
			0.7298$\pm$0.0035 &
			\multicolumn{1}{c}{0.7347$\pm$0.0055} &
			\multicolumn{1}{c}{0.7299$\pm$0.0105} &
			0.7344$\pm$0.0056 &
			\multicolumn{1}{c}{0.7398$\pm$0.0052} &
			\multicolumn{1}{c}{0.7349$\pm$0.0034} &
			0.7396$\pm$0.0051 \\ \hline
		\end{tabular*}%
	}
\end{table*}
\begin{table*}[bht!]
	\centering
	\tiny
	\caption{Performance comparison over Fact-checked News}
	\label{tab:results_fact_checked}
	\resizebox{\textwidth}{!}{%
		\begin{tabular*}{\textwidth}{@{\extracolsep{\fill}}c ccc ccc ccc}
			\hline
			\multirow{2}{*}{\textbf{method}} &
			\multicolumn{3}{c}{\textbf{10\%}} &
			\multicolumn{3}{c}{\textbf{20\%}} &
			\multicolumn{3}{c}{\textbf{30\%}} \\ \cline{2-10} 
			&
			\multicolumn{1}{c}{\textbf{Acc}} &
			\multicolumn{1}{c}{\textbf{Intereset-F1}} &
			\textbf{Macro-F1} &
			\multicolumn{1}{c}{\textbf{Acc}} &
			\multicolumn{1}{c}{\textbf{Intereset-F1}} &
			\textbf{Macro-F1} &
			\multicolumn{1}{c}{\textbf{Acc}} &
			\multicolumn{1}{c}{\textbf{Intereset-F1}} &
			\textbf{Macro-F1} \\ \hline
			LOSS-GAT &
			\multicolumn{1}{c}{\textbf{0.9163$\pm$0.0019}} &
			\multicolumn{1}{c}{\textbf{0.9127$\pm$0.0022}} &
			\textbf{0.9161$\pm$0.0019} &
			\multicolumn{1}{c}{\textbf{0.9186$\pm$0.0016}} &
			\multicolumn{1}{c}{\textbf{0.9101$\pm$0.0018}} &
			\textbf{0.9179$\pm$0.0016} &
			\multicolumn{1}{c}{\textbf{0.9208$\pm$0.0028}} &
			\multicolumn{1}{c}{\textbf{0.9055$\pm$0.0036}} &
			\textbf{0.9186$\pm$0.0029} \\ \hline
			PULP-LPHN &
			\multicolumn{1}{c}{0.776$\pm$0.0083} &
			\multicolumn{1}{c}{0.7408$\pm$0.0115} &
			0.7717$\pm$0.0088 &
			\multicolumn{1}{c}{0.8094$\pm$0.0052} &
			\multicolumn{1}{c}{0.7749$\pm$0.0091} &
			0.8048$\pm$0.0059 &
			\multicolumn{1}{c}{0.8328$\pm$0.0098} &
			\multicolumn{1}{c}{0.7939$\pm$0.0122} &
			0.8266$\pm$0.0101 \\ \hline
			OCSVM &
			\multicolumn{1}{c}{0.6576$\pm$0.0132} &
			\multicolumn{1}{c}{0.5693$\pm$0.0171} &
			0.6423$\pm$0.0125 &
			\multicolumn{1}{c}{0.6713$\pm$0.008} &
			\multicolumn{1}{c}{0.5669$\pm$0.0146} &
			0.6509$\pm$0.0082 &
			\multicolumn{1}{c}{0.6819$\pm$0.0051} &
			\multicolumn{1}{c}{0.5761$\pm$0.0091} &
			0.6606$\pm$0.0033 \\ \hline
			OCRF &
			\multicolumn{1}{c}{0.6553$\pm$0.0063} &
			\multicolumn{1}{c}{0.5909$\pm$0.02} &
			0.6461$\pm$0.008 &
			\multicolumn{1}{c}{0.6682$\pm$0.0079} &
			\multicolumn{1}{c}{0.5811$\pm$0.0252} &
			0.6526$\pm$0.0098 &
			\multicolumn{1}{c}{0.6876$\pm$0.0094} &
			\multicolumn{1}{c}{0.5726$\pm$0.017} &
			0.6632$\pm$0.0108 \\ \hline
			KNND &
			\multicolumn{1}{c}{0.5696$\pm$0.0026} &
			\multicolumn{1}{c}{0.563$\pm$0.0038} &
			0.5695$\pm$0.0027 &
			\multicolumn{1}{c}{0.5703$\pm$0.0027} &
			\multicolumn{1}{c}{0.549$\pm$0.0042} &
			0.5694$\pm$0.0028 &
			\multicolumn{1}{c}{0.5712$\pm$0.0027} &
			\multicolumn{1}{c}{0.5322$\pm$0.0044} &
			0.5682$\pm$0.0029 \\ \hline
			KMeans &
			\multicolumn{1}{c}{0.5449$\pm$0.008} &
			\multicolumn{1}{c}{0.4495$\pm$0.0208} &
			0.5301$\pm$0.003 &
			\multicolumn{1}{c}{0.5573$\pm$0.0089} &
			\multicolumn{1}{c}{0.4279$\pm$0.0177} &
			0.533$\pm$0.0048 &
			\multicolumn{1}{c}{0.5768$\pm$0.0059} &
			\multicolumn{1}{c}{0.4046$\pm$0.0047} &
			0.5381$\pm$0.0029 \\ \hline
			BLDT &
			\multicolumn{1}{c}{0.837$\pm$0.0079} &
			\multicolumn{1}{c}{0.8407$\pm$0.0077} &
			0.8368$\pm$0.008 &
			\multicolumn{1}{c}{0.8479$\pm$0.012} &
			\multicolumn{1}{c}{0.8495$\pm$0.0124} &
			0.8479$\pm$0.012 &
			\multicolumn{1}{c}{0.8666$\pm$0.0068} &
			\multicolumn{1}{c}{0.8683$\pm$0.0067} &
			0.8665$\pm$0.0067 \\ \hline
			BLRF &
			\multicolumn{1}{c}{0.8954$\pm$0.0043} &
			\multicolumn{1}{c}{0.8966$\pm$0.0054} &
			0.8953$\pm$0.0043 &
			\multicolumn{1}{c}{0.9024$\pm$0.0061} &
			\multicolumn{1}{c}{0.9036$\pm$0.0063} &
			0.9024$\pm$0.0061 &
			\multicolumn{1}{c}{0.9037$\pm$0.0043} &
			\multicolumn{1}{c}{0.9049$\pm$0.004} &
			0.9036$\pm$0.0043 \\ \hline
		\end{tabular*}%
	}
\end{table*}
\begin{table*}[bht!]
	\centering
	\tiny
	\caption{Performance comparison over Fake News Detection}
	\label{tab:results_fn_detection}
	\resizebox{\textwidth}{!}{%
		\begin{tabular*}{\textwidth}{@{\extracolsep{\fill}}c ccc ccc ccc}
			\hline
			\multirow{2}{*}{\textbf{method}} &
			\multicolumn{3}{c}{\textbf{10\%}} &
			\multicolumn{3}{c}{\textbf{20\%}} &
			\multicolumn{3}{c}{\textbf{30\%}} \\ \cline{2-10} 
			&
			\multicolumn{1}{c}{\textbf{Acc}} &
			\multicolumn{1}{c}{\textbf{Intereset-F1}} &
			\textbf{Macro-F1} &
			\multicolumn{1}{c}{\textbf{Acc}} &
			\multicolumn{1}{c}{\textbf{Intereset-F1}} &
			\textbf{Macro-F1} &
			\multicolumn{1}{c}{\textbf{Acc}} &
			\multicolumn{1}{c}{\textbf{Intereset-F1}} &
			\textbf{Macro-F1} \\ \hline
			LOSS-GAT &
			\multicolumn{1}{c}{\textbf{0.8331$\pm$0.0117}} &
			\multicolumn{1}{c}{\textbf{0.8378$\pm$0.0125}} &
			\textbf{0.8329$\pm$0.0117} &
			\multicolumn{1}{c}{\textbf{0.8546$\pm$0.0072}} &
			\multicolumn{1}{c}{\textbf{0.8508$\pm$0.0094}} &
			\textbf{0.8544$\pm$0.0073} &
			\multicolumn{1}{c}{\textbf{0.8685$\pm$0.0027}} &
			\multicolumn{1}{c}{\textbf{0.8574$\pm$0.0035}} &
			\textbf{0.8677$\pm$0.0027} \\ \hline
			PULP-LPHN &
			\multicolumn{1}{c}{0.6873$\pm$0.0108} &
			\multicolumn{1}{c}{0.6703$\pm$0.0123} &
			0.6862$\pm$0.0105 &
			\multicolumn{1}{c}{0.7442$\pm$0.0062} &
			\multicolumn{1}{c}{0.7332$\pm$0.0069} &
			0.7437$\pm$0.0061 &
			\multicolumn{1}{c}{0.7943$\pm$0.0058} &
			\multicolumn{1}{c}{0.7708$\pm$0.0074} &
			0.7921$\pm$0.0059 \\ \hline
			OCSVM &
			\multicolumn{1}{c}{0.675$\pm$0.0054} &
			\multicolumn{1}{c}{0.6187$\pm$0.0195} &
			0.6675$\pm$0.0082 &
			\multicolumn{1}{c}{0.678$\pm$0.0033} &
			\multicolumn{1}{c}{0.5889$\pm$0.0158} &
			0.662$\pm$0.0065 &
			\multicolumn{1}{c}{0.6918$\pm$0.0045} &
			\multicolumn{1}{c}{0.5862$\pm$0.0129} &
			0.6702$\pm$0.0059 \\ \hline
			OCRF &
			\multicolumn{1}{c}{0.6722$\pm$0.0035} &
			\multicolumn{1}{c}{0.651$\pm$0.0107} &
			0.6707$\pm$0.0034 &
			\multicolumn{1}{c}{0.6737$\pm$0.0042} &
			\multicolumn{1}{c}{0.6268$\pm$0.0133} &
			0.6682$\pm$0.0049 &
			\multicolumn{1}{c}{0.682$\pm$0.006} &
			\multicolumn{1}{c}{0.6138$\pm$0.0116} &
			0.6715$\pm$0.0043 \\ \hline
			KNND &
			\multicolumn{1}{c}{0.483$\pm$0.0033} &
			\multicolumn{1}{c}{0.4703$\pm$0.0249} &
			0.4815$\pm$0.0023 &
			\multicolumn{1}{c}{0.4926$\pm$0.0058} &
			\multicolumn{1}{c}{0.414$\pm$0.0162} &
			0.4828$\pm$0.0034 &
			\multicolumn{1}{c}{0.5014$\pm$0.0048} &
			\multicolumn{1}{c}{0.396$\pm$0.0138} &
			0.4855$\pm$0.0029 \\ \hline
			KMeans &
			\multicolumn{1}{c}{0.6315$\pm$0.0024} &
			\multicolumn{1}{c}{0.5959$\pm$0.0036} &
			0.6286$\pm$0.0025 &
			\multicolumn{1}{c}{0.6371$\pm$0.0024} &
			\multicolumn{1}{c}{0.5853$\pm$0.0038} &
			0.6314$\pm$0.0026 &
			\multicolumn{1}{c}{0.6434$\pm$0.0027} &
			\multicolumn{1}{c}{0.572$\pm$0.0046} &
			0.6332$\pm$0.0031 \\ \hline
			BLDT &
			\multicolumn{1}{c}{0.775$\pm$0.0065} &
			\multicolumn{1}{c}{0.7871$\pm$0.0081} &
			0.7742$\pm$0.0063 &
			\multicolumn{1}{c}{0.7945$\pm$0.0057} &
			\multicolumn{1}{c}{0.8091$\pm$0.0065} &
			0.7932$\pm$0.0057 &
			\multicolumn{1}{c}{0.8055$\pm$0.0049} &
			\multicolumn{1}{c}{0.8164$\pm$0.0044} &
			0.8047$\pm$0.0051 \\ \hline
			BLRF &
			\multicolumn{1}{c}{0.8141$\pm$0.0084} &
			\multicolumn{1}{c}{0.824$\pm$0.0107} &
			0.8134$\pm$0.0081 &
			\multicolumn{1}{c}{0.8287$\pm$0.006} &
			\multicolumn{1}{c}{0.8427$\pm$0.0064} &
			0.8273$\pm$0.0061 &
			\multicolumn{1}{c}{0.8248$\pm$0.0079} &
			\multicolumn{1}{c}{0.8366$\pm$0.009} &
			0.8238$\pm$0.0077 \\ \hline
		\end{tabular*}%
	}
\end{table*}
\begin{table*}[bht!]
	\centering
	\tiny
	\caption{Performance comparison over Fake News Data.}
	\label{tab:results_fn_data}
	\resizebox{\textwidth}{!}{%
		\begin{tabular*}{\textwidth}{@{\extracolsep{\fill}}c ccc ccc ccc}
			\hline
			\multirow{2}{*}{\textbf{method}} &
			\multicolumn{3}{c}{\textbf{10\%}} &
			\multicolumn{3}{c}{\textbf{20\%}} &
			\multicolumn{3}{c}{\textbf{30\%}} \\ \cline{2-10} 
			&
			\multicolumn{1}{c}{\textbf{Acc}} &
			\multicolumn{1}{c}{\textbf{Intereset\_F1}} &
			\textbf{Macro\_F1\_10} &
			\multicolumn{1}{c}{\textbf{Acc}} &
			\multicolumn{1}{c}{\textbf{Intereset\_F1}} &
			\textbf{Macro\_F1} &
			\multicolumn{1}{c}{\textbf{Acc}} &
			\multicolumn{1}{c}{\textbf{Intereset\_F1}} &
			\textbf{Macro\_F1} \\ \hline
			Proposed &
			\multicolumn{1}{c}{\textbf{0.8218$\pm$0.0102}} &
			\multicolumn{1}{c}{\textbf{0.83$\pm$0.0091}} &
			\textbf{0.8213$\pm$0.0105} &
			\multicolumn{1}{c}{\textbf{0.8028$\pm$0.0165}} &
			\multicolumn{1}{c}{\textbf{0.813$\pm$0.0117}} &
			\textbf{0.802$\pm$0.0171} &
			\multicolumn{1}{c}{\textbf{0.8362$\pm$0.0085}} &
			\multicolumn{1}{c}{\textbf{0.8398$\pm$0.0094}} &
			\textbf{0.8361$\pm$0.0085} \\ \hline
			
			PULP-LPHN &
			\multicolumn{1}{c}{0.5008$\pm$0.0054} &
			\multicolumn{1}{c}{0.3901$\pm$0.011} &
			0.4837$\pm$0.0065 &
			\multicolumn{1}{c}{0.4991$\pm$0.008} &
			\multicolumn{1}{c}{0.4052$\pm$0.0076} &
			0.4863$\pm$0.0078 &
			\multicolumn{1}{c}{0.5021$\pm$0.0033} &
			\multicolumn{1}{c}{0.4032$\pm$0.0037} &
			0.488$\pm$0.0027 \\ \hline
			OCSVM &
			\multicolumn{1}{c}{0.6278$\pm$0.0009} &
			\multicolumn{1}{c}{0.6088$\pm$0.001} &
			0.6269$\pm$0.0009 &
			\multicolumn{1}{c}{0.6265$\pm$0.0018} &
			\multicolumn{1}{c}{0.6137$\pm$0.0035} &
			0.6261$\pm$0.0018 &
			\multicolumn{1}{c}{0.6271$\pm$0.0015} &
			\multicolumn{1}{c}{0.6154$\pm$0.0014} &
			0.6268$\pm$0.0014 \\ \hline
			OCRF &
			\multicolumn{1}{c}{0.6276$\pm$0.0021} &
			\multicolumn{1}{c}{0.6255$\pm$0.004} &
			0.6275$\pm$0.0021 &
			\multicolumn{1}{c}{0.6267$\pm$0.001} &
			\multicolumn{1}{c}{0.623$\pm$0.0033} &
			0.6266$\pm$0.001 &
			\multicolumn{1}{c}{0.6286$\pm$0.0021} &
			\multicolumn{1}{c}{0.623$\pm$0.002} &
			0.6284$\pm$0.002 \\ \hline
			KNND &
			\multicolumn{1}{c}{0.6222$\pm$0.0008} &
			\multicolumn{1}{c}{0.6236$\pm$0.0133} &
			0.6217$\pm$0.0008 &
			\multicolumn{1}{c}{0.6219$\pm$0.001} &
			\multicolumn{1}{c}{0.6081$\pm$0.0031} &
			0.6214$\pm$0.0009 &
			\multicolumn{1}{c}{0.6223$\pm$0.0009} &
			\multicolumn{1}{c}{0.6277$\pm$0.0121} &
			0.6218$\pm$0.0009 \\ \hline
			KMeans &
			\multicolumn{1}{c}{0.6023$\pm$0.0013} &
			\multicolumn{1}{c}{0.5746$\pm$0.0014} &
			0.6006$\pm$0.0013 &
			\multicolumn{1}{c}{0.6023$\pm$0.0014} &
			\multicolumn{1}{c}{0.5743$\pm$0.0015} &
			0.6006$\pm$0.0014 &
			\multicolumn{1}{c}{0.6029$\pm$0.0006} &
			\multicolumn{1}{c}{0.5741$\pm$0.0011} &
			0.6011$\pm$0.0007 \\ \hline
			BLDT &
			\multicolumn{1}{c}{0.7284$\pm$0.0045} &
			\multicolumn{1}{c}{0.7325$\pm$0.0076} &
			0.7281$\pm$0.0047 &
			\multicolumn{1}{c}{0.7445$\pm$0.0025} &
			\multicolumn{1}{c}{0.7511$\pm$0.0051} &
			0.7442$\pm$0.0023 &
			\multicolumn{1}{c}{0.7495$\pm$0.0025} &
			\multicolumn{1}{c}{0.7534$\pm$0.0036} &
			0.7494$\pm$0.0025 \\ \hline
			BLRF &
			\multicolumn{1}{c}{0.7928$\pm$0.0037} &
			\multicolumn{1}{c}{0.7947$\pm$0.005} &
			0.7927$\pm$0.0037 &
			\multicolumn{1}{c}{0.8005$\pm$0.005} &
			\multicolumn{1}{c}{0.8035$\pm$0.0077} &
			0.8003$\pm$0.005 &
			\multicolumn{1}{c}{0.7989$\pm$0.0057} &
			\multicolumn{1}{c}{0.8051$\pm$0.004} &
			0.7987$\pm$0.0058 \\ \hline
		\end{tabular*}%
	}
\end{table*}
\subsubsection{	Q1: To what extent does the two-step label propagation process contribute to enhancing the model's performance?}
\label{sec:q1}
\begin{figure}[!t]
	\centering
	\subfigure[]{\includegraphics[width=0.23\textwidth]{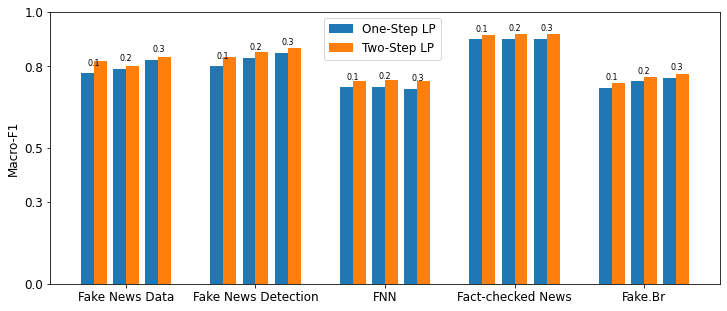}}
	\subfigure[]{\includegraphics[width=0.23\textwidth]{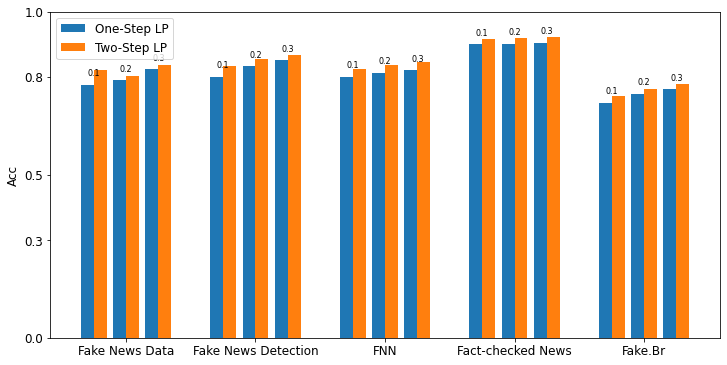}}
	\caption{Macro-F1 (a) and Accuracy (b) comparision of two-step label propagation and one-step label propagation.
		0.1, 0.2, and 0.3 represent the amount of labeled data used to train the algorithms.}
	\label{fig:lp_f1_acc}
\end{figure}

To address this question, we conduct a semi-supervised classification after the first label propagation step to predict the final labels using GNNs.
We then compare these results with the outcomes obtained from our proposed method,
assessing the effectiveness of two-step label propagation.
For each dataset, in \autoref{fig:lp_f1_acc} we present the charts depicting the macro-F1 and accuracy values, for all three scenarios: 10\%, 20\% and 30\% labeled fake news.
As evident, across all the datasets and in all three scenarios, the inclusion of the second step in label propagation consistently results in improved model performance.
To conduct a closer examination of the effect of two-step label propagation,
we present in \autoref{tab:lp_improvement} the subtraction of macro-F1 (accuracy) values between the two-step method and the one-step method. 
In this table, the "Acc-improvement" and "Macro-F1-improvement" columns denote the subtraction of accuracy and macro-F1 values between the two-step and one-step methods, respectively.
Inclusion of the second label propagation step leads to improvements ranging from +1\% to +3\%.
The most significant improvement is observed on the Fake News Detection dataset in the 10\% labeled scenario, while the smallest enhancement is seen on the Fake.Br dataset in the 30\% labeled scenario.
These results imply that the incorporation of a second step in label propagation, along with an initial classifier to identify pseudo-labels, can yield enhanced classification outcomes.
\begin{table*}[!t]
	\centering
	\caption{Differences between macro-F1 and accuracy values of the two-step and the one-step label propagation methods.}
	\label{tab:lp_improvement}	
	\begin{tabular}{c c c c}
		\hline
		\textbf{Dataset} & \textbf{\% labeled} & \textbf{Macro-F1-improvement} & \textbf{Acc-improvement} \\ \hline
		
		\multirow{3}{*}{FNN}         & 10\% & 0.0247 & 0.0254 \\ \cline{2-4} 
		& 20\% & 0.0244 & 0.0241 \\ \cline{2-4} 
		& 30\% & 0.0276 & 0.0244 \\ \hline
		
		\multirow{3}{*}{Fake.Br}      & 10\% & 0.0201 & 0.0209 \\ \cline{2-4} 
		& 20\% & 0.0157 & 0.0164 \\ \cline{2-4} 
		& 30\% & 0.014  & 0.0149 \\ \hline
		
		\multirow{3}{*}{Fact-checked News} & 10\% & 0.0169 & 0.0168 \\ \cline{2-4} 
		& 20\% & 0.0166 & 0.0164 \\ \cline{2-4} 
		& 30\% & 0.0167 & 0.0166 \\ \hline
		
		\multirow{3}{*}{Fake News Detection} & 10\% & 0.0325 & 0.0323 \\ \cline{2-4} 
		& 20\% & 0.0224 & 0.0224 \\ \cline{2-4} 
		& 30\% & 0.0169 & 0.017  \\ \hline
		
		\multirow{3}{*}{Fake News Data} & 10\% & 0.0469 & 0.0454 \\ \cline{2-4} 
		& 20\% & 0.0120 & 0.0131 \\ \cline{2-4} 
		& 30\% & 0.0135 & 0.0122  \\ \hline
	\end{tabular}%
\end{table*}
\subsubsection{	Q2: Does the quantity of labeled data significantly influence the final results?}
\label{sec:q2}
\begin{figure*}[!t]
	\centering
	\subfigure[]{\includegraphics[width=0.18\textwidth]{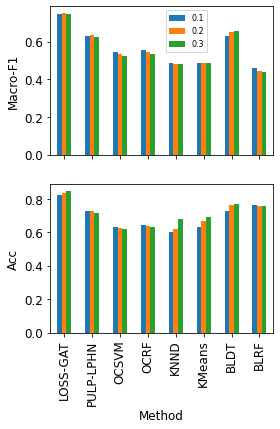}}
	\subfigure[]{\includegraphics[width=0.18\textwidth]{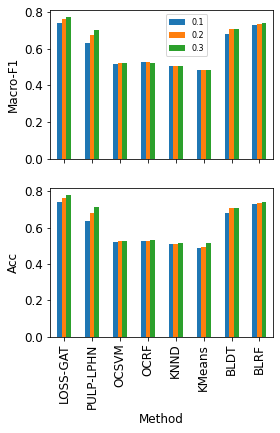}}
	\subfigure[]{\includegraphics[width=0.18\textwidth]{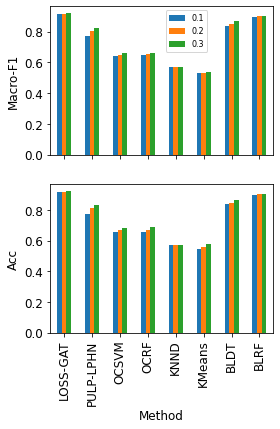}}
	\subfigure[]{\includegraphics[width=0.18\textwidth]{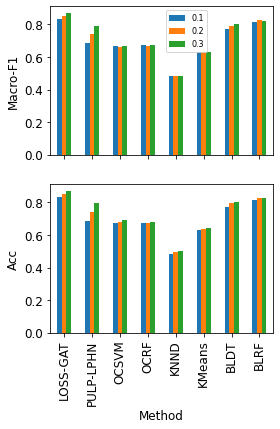}}
	\subfigure[]{\includegraphics[width=0.18\textwidth]{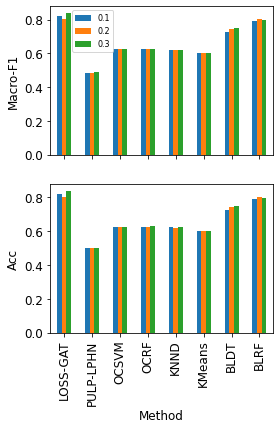}}
	
	\caption{Macro-F1 and accuracy comparision, for different amounts of labeled fake news.
		0.1, 0.2, and 0.3 represent the ratio of labeled data used to train the algorithms. (a): FNN, (b): Fake.Br, (c): Fact-checked News,
		(d): Fake News Detection, (e): Fake News Data.}	
	\label{fig:amount_labeld}
\end{figure*}

As observed in \autoref{fig:amount_labeld}, it can be inferred that in overall,
changing the quantity of labeled fake news in the training data does not yield a substantial difference in the final results. As shown in \autoref{sec:q1}, some algorithms such as OCRF and OCSVM are able to achieve the maximum values of macro-F1 or accuracy with only 10\% of labeled fake news. Hence, it can be deduced that OCL-based methods can achieve commendable outcomes using solely quantitatively labeled data from the class of interest.
\subsubsection{	Q3: Has the OCL approach demonstrated superior performance compared to binary labeled methods?}
\label{sec:q3}
According to \autoref{tab:results_fnn}, \autoref{tab:results_fakebr}, \autoref{tab:results_fact_checked}, \autoref{tab:results_fn_detection}, and \autoref{tab:results_fn_data}, it is evident that OCL methods do not invariably yield superior results. 
For instance, a performance comparison between OCL methods such as OCSVM, OCRF with the BLDT method (a widely-used binary-labeled classification algorithm), reveals that these OCL methods, which exclusively utilize data with fake news labels, do not consistently outperform the BLDT algorithm.
Both the PULP-LPHN method and our proposed approach incorporate a blend of label propagation within the OCL framework. 
The outcomes demonstrate that the PULP-LPHN method outperforms OCL methods and exhibits results that are on par with or even superior to the BLDT method. 
Ultimately, our proposed method consistently achieves higher results than binary classification algorithms, across all scenarios under investigation.
These results underscore that by fusing the OCL approach with label propagation and implementing a well-devised strategy, it is possible to achieve superior performance compared to binary labeled methods, even with access to a limited set of positive class data (in this case, fake news). This effectively diminishes the model's reliance on a substantial volume of labeled data from both classes.
\subsubsection{	Q4: Does LOSS-GAT outperform other baseline models for fake news detection?}
\label{sec:q4}
In \autoref{tab:avg_ranking},
we  present the mean average ranking values for the baselines and our proposed model, considering macro-F1 and accuracy.
10\%, 20\% and 30\% show the percentages of labeled data used to train models.
The best performances are highlighted in bold.
As seen in the table, our proposed method achieves the best mean average ranking among the examined models.
Binary labeled methods come second.
The PULP-LPHN method performs better than the other OCL methods, but its performance is not as good as binary labeled methods.
Therefore, as the results presented in \autoref{tab:results_fnn}, \autoref{tab:results_fakebr}, \autoref{tab:results_fact_checked},  \autoref{tab:results_fn_detection} and  \autoref{tab:results_fn_data} also confirm, our proposed method, which uses only a limited amount of data labeled as fake news, by leveraging the power of GNNs and proposing a two-step label propagation algorithm,  performs even better than the binary labeled methods.
\begin{table*}[bht!]
	\centering
	\caption{Average ranking values of the macro-F1 and accuracy results, for the 10\%, 20\% and 30\% of labeled data scenarios. The last column presents the mean of the average rankings.}
	\label{tab:avg_ranking}	
	\begin{tabular}{c cccc cccc}
		\hline
		\multirow{2}{*}{\textbf{method}} &
		\multicolumn{4}{c}{\textbf{ACC}} &
		\multicolumn{4}{c}{\textbf{Macro-F1}} \\ \cline{2-9} 
		&
		\multicolumn{1}{c}{\textbf{10\%}} &
		\multicolumn{1}{c}{\textbf{20\%}} &
		\multicolumn{1}{c}{\textbf{30\%}} &
		\textbf{Avg} &
		\multicolumn{1}{c}{\textbf{10\%}} &
		\multicolumn{1}{c}{\textbf{20\%}} &
		\multicolumn{1}{c}{\textbf{30\%}} &
		\textbf{Avg} \\ \hline
		LOSS-GAT &
		\multicolumn{1}{c}{\textbf{1}} &
		\multicolumn{1}{c}{\textbf{1}} &
		\multicolumn{1}{c}{\textbf{1}} &
		\textbf{1} &
		\multicolumn{1}{c}{\textbf{1}} &
		\multicolumn{1}{c}{\textbf{1}} &
		\multicolumn{1}{c}{\textbf{1}} &
		\textbf{1} \\ \hline
		PULP-LPHN &
		\multicolumn{1}{c}{4.6} &
		\multicolumn{1}{c}{4.8} &
		\multicolumn{1}{c}{4.6} &
		4.7 &
		\multicolumn{1}{c}{4.6} &
		\multicolumn{1}{c}{4.6} &
		\multicolumn{1}{c}{4.6} &
		4.6 \\ \hline
		OCSVM &
		\multicolumn{1}{c}{5.2} &
		\multicolumn{1}{c}{5.6} &
		\multicolumn{1}{c}{6} &
		5.6 &
		\multicolumn{1}{c}{5.6} &
		\multicolumn{1}{c}{5.6} &
		\multicolumn{1}{c}{5.6} &
		5.6 \\ \hline
		OCRF &
		\multicolumn{1}{c}{5.4} &
		\multicolumn{1}{c}{5.4} &
		\multicolumn{1}{c}{5.4} &
		5.4 &
		\multicolumn{1}{c}{4.6} &
		\multicolumn{1}{c}{4.6} &
		\multicolumn{1}{c}{4.6} &
		4.6 \\ \hline
		KNND &
		\multicolumn{1}{c}{7.2} &
		\multicolumn{1}{c}{7.2} &
		\multicolumn{1}{c}{7} &
		7.1 &
		\multicolumn{1}{c}{7} &
		\multicolumn{1}{c}{7} &
		\multicolumn{1}{c}{7} &
		7 \\ \hline
		KMeans &
		\multicolumn{1}{c}{7.4} &
		\multicolumn{1}{c}{7} &
		\multicolumn{1}{c}{6.8} &
		7.1 &
		\multicolumn{1}{c}{7.2} &
		\multicolumn{1}{c}{7.2} &
		\multicolumn{1}{c}{7.2} &
		7.2 \\ \hline
		BLDT &
		\multicolumn{1}{c}{3.2} &
		\multicolumn{1}{c}{2.8} &
		\multicolumn{1}{c}{3} &
		3 &
		\multicolumn{1}{c}{2.8} &
		\multicolumn{1}{c}{2.8} &
		\multicolumn{1}{c}{2.8} &
		2.8 \\ \hline
		BLRF &
		\multicolumn{1}{c}{2} &
		\multicolumn{1}{c}{2.2} &
		\multicolumn{1}{c}{2.2} &
		2.1 &
		\multicolumn{1}{c}{3.2} &
		\multicolumn{1}{c}{3.2} &
		\multicolumn{1}{c}{3.2} &
		3.2 \\ \hline
	\end{tabular}%
\end{table*}
\section{Conclusion}
\label{sec:conclusion}
In this paper, we presented a new method to detect fake news based on a One-Class Learning approach, using label propagation methods and leveraging the power of GNNs in semi-supervised learning, with the aim of minimizing the efforts required for data labeling.
Our proposed method extracts the textual and structural information from the unlabeled data, by utilizing only a small number of data labeled as fake news. 
It constructs a similarity graph and applies a two-step label propagation algorithm.
The two-level label propagation, along with applying a structural augmentation on the graph and inducing randomness in the local neighborhood of the nodes,
are among the techniques that our proposed model uses to classify news as fake or real.
We evaluated our proposed model over five real-world datasets (balanced and imbalanced),
in the English and Portuguese languages.
We compared it against a number of One-Class Learning approaches and a binary labeled method.
Our experiments demonstrated the effectiveness of our proposed method in enhancing the detection performance of fake news, while requiring a smaller training data  (only a small amount of data labeled as fake news).

\printbibliography 


\end{document}